\documentclass[journal]{IEEEtran}

\usepackage{cite}
\usepackage[pdftex]{graphicx} 
\usepackage{amsmath}
\usepackage{algorithmic}

\usepackage{array}

\usepackage[caption=false,font=footnotesize]{subfig}

\usepackage{xcolor}

\usepackage{soul}

\renewcommand{\ul}{}

\begin{document}
%
\title{Data-driven Modeling of Mach-Zehnder Interferometer-based Optical Matrix Multipliers}
%
%
%

\author{Ali~Cem,~\IEEEmembership{Student Member,~IEEE,}
        Siqi~Yan,
        Yunhong~Ding,
        Darko~Zibar,
        and~Francesco~Da~Ros,~\IEEEmembership{Senior Member,~IEEE}
        
\thanks{Manuscript received October 17, 2022; revised March 03, 2023.}
\thanks{Ali Cem, Yunhong Ding, Darko Zibar and Francesco Da Ros are with the Department of Electrical and Photonics Engineering, Technical University of Denmark, 2800 Lyngby, Denmark (e-mail: alice@dtu.dk, yudin@dtu.dk, dazi@dtu.dk, fdro@dtu.dk).}
\thanks{Siqi Yan is with the School of Optical and Electronic Information, Huazhong University of Science and Technology, 430074 Wuhan, China (e-mail: siqya@mail.hust.edu.cn).}
\thanks{Online availability.}}

\markboth{Journal of Lightwave Technology,~Vol.~??, No.~?, Month~Year}%
{Cem \MakeLowercase{\textit{et al.}}: Paper Title}

\maketitle

\begin{abstract} 
Photonic integrated circuits are facilitating the development of optical neural networks, which have the potential to be both faster and more energy efficient than their electronic counterparts since optical signals are especially well-suited for implementing matrix multiplications. However, accurate programming of photonic chips for optical matrix multiplication remains a difficult challenge. Here, we describe both simple analytical models and data-driven models for offline training of optical matrix multipliers. We train and evaluate the models using experimental data obtained from a fabricated chip featuring a Mach-Zehnder interferometer mesh implementing 3-by-3 matrix multiplication. The neural network-based models outperform the simple physics-based models in terms of prediction error. Furthermore, the neural network models are also able to predict the spectral variations in the matrix weights for up to 100 frequency channels covering the C-band. The use of neural network models for programming the chip for optical matrix multiplication yields increased performance on multiple machine learning tasks.  
\end{abstract}


\IEEEpeerreviewmaketitle

\section{Introduction}

\IEEEPARstart{T}{he} field of artificial intelligence has undergone major changes in the last decade, which can be attributed to the success of deep learning when working with large datasets and advances in computational hardware \cite{Shastri2021}. However, as the post-Moore era draws near, conventional computers are not projected to keep up with the increasing demands \cite{Huang2022}. Neuromorphic engineering promises to achieve new hardware architectures that match the distributed nature of machine learning (ML) algorithms, leading the way towards faster and more sustainable computation \cite{Shastri2021,Huang2022}. Physical implementations of neural networks (NNs) require a very large number of parallel interconnections, which is difficult to realize efficiently using electronics and metal wiring \cite{Nahmias2019}. The trade-off between interconnectivity and signal bandwidth limits the processing speed of neuromorphic electronics, which has lead to the investigation of non-CMOS platforms for computation.

In particular, photonic integrated circuits (PICs) are very suitable for NN implementations, as optical waveguides have significantly higher bandwidths and lower losses compared to metal wires that additionally need to charge or discharge electrically when changing states, increasing energy consumption \cite{Zhou2022}. This is critical when implementing the massively-connected linear layers of NNs. A large variety of architectures for optical NNs have been proposed, including but not limited to Mach-Zehnder interferometer (MZI) meshes \cite{Shen2017,Marinis2021}, microring weight banks \cite{Tait2017,Ashtiani2022,Zhang2022}, SOA-based architectures \cite{Shi2022}, photonic crossbar arrays \cite{Youngblood2022,Vlieg2022}, and coherent optical neural networks enabled by wavelength division multiplexing \cite{Totovic2022}.


MZI meshes are used for optical matrix multiplication (OMM) to implement the linear layers, which can then be combined with all-optical or optoelectronic nonlinear layers \cite{Campo2022,Williamson2019} to realize feedforward optical neural networks. Over the last years, different mesh topologies were used to implement linear operations \cite{Marinis2021}. In most cases, the linear weights are tuned using a set of voltages applied to the thermo-optic phase shifters of the MZIs \cite{Bogaerts2020}. To be able to program a MZI mesh to accurately implement a desired linear operation, it is desirable to have a model describing the matrix weights as a function of the heater voltages. One way of obtaining such a model, commonly referred to as the offline (\textit{in silico}) training of the chip, is to rely on the well-known analytical expressions describing the transmission through a MZI \cite{Perez2017}. However, simple physics-based approaches result in high modeling errors for a fabricated chip due to fabrication errors and additional effects that are challenging to model accurately, such as deterministic thermal crosstalk between the heaters \cite{Perez2017,Fang2019,Bandyopadhyay2021}. Due to these practical limitations of the simple physics-based models, a variety of techniques have been developed to program PICs accurately for OMM \cite{Miller2013,Fang2019,Bandyopadhyay2021,Pai2020,Semenova2022,Paolini2022}. While such procedures can be used to correct inaccuracies due to fabrication errors, they cannot tackle additional causes of inaccuracy such as thermal crosstalk. A Bayesian training scheme has been proposed in \cite{Sarantoglou2022} that manages to reduce the impact of low thermal crosstalk on performance, but does not alleviate it fully even though thermal crosstalk is mostly deterministic. Despite the advances in crosstalk-minimizing PIC design \cite{Milanizadeh2020}, thermal crosstalk remains to be a limiting factor for scaling up the sizes of OMMs, as its significance increases with the number of MZIs per PIC footprint. Finally, instead of training the chip offline, PICs can be programmed accurately through the use of online (\textit{in situ}) training procedures \cite{Cong2019,Zhang2021,Gu2022}, but this approach requires re-optimizing the heater voltages whenever a new linear operation needs to be implemented. This in turn limits the potential use cases, as well as potentially requiring additional hardware since monitor ports may need to be added \cite{Wright2022}.

It has recently been shown that ML methods can be used to model MZI meshes offline even in the presence of fabrication tolerances \cite{Cem2022,Feng2022}. Given enough measurement data for training, NN models can accurately predict the implemented weights for heater voltages. This data-driven modeling approach can also be used to simultaneously model the spectral response of the PIC for up to 100 wavelengths covering the entire C-band \cite{CemIPC2022}. In contrast to previous single-wavelength models, models that can handle multiple wavelengths can enable applications involving wavelength multiplexing or they can be used to optimize the spectral shape of the input and the output signals to the PIC.

In this work, we extend upon our previous work on data-driven modeling of MZI meshes in \cite{Cem2022} and \cite{CemIPC2022} by analyzing the performance of a variety of NN architectures for modeling of a fabricated PIC in detail. In addition to comparing the data-driven model to physics-based analytical models for single-wavelength modeling, we also compare and contrast how different data-driven modeling approaches perform for multiple-wavelength modeling. 

The paper is structured as follows: In Chapter 2, we explain the use of MZI meshes for OMM in detail and describe our experimental setup. In Chapter 3, we describe and justify different modeling approaches. Finally in Chapter 4, we present our experimental results for the models and compare their performance for single-wavelength and multiple-wavelength modeling. In addition to their modeling accuracies, we also consider the impact of the training set size on the modeling performance and relate the final modeling errors to the performance on two ML tasks where the OMM model is used to implement a linear layer.

\emph{Notation:} Bold letters indicate vectors and matrices/tensors, where lowercase letters are used for the former and uppercase letters for the latter. Superscripts in parentheses such as the $l$ in $\mathbf{W}^{(l)}$ denote the index for a particular measurement. Subscripts are used for indices of matrix or vector entries and they are separated by commas for multi-dimensional quantities as in $w^{(l)}_{i,j,k}$.

\section{OMM Using MZI Meshes}

\subsection{MZI Meshes for OMM: Working Principle}

Each hidden layer of a feedforward optical NN consists of a linear layer that can be represented by a matrix-vector multiplication and an element-wise nonlinear activation layer. In this work, we will only focus on the linear layers implemented using MZI meshes. In order to implement multiplication by any real-valued $M\times N$ matrix $\mathbf{A}$, the singular value decomposition of $\mathbf{A}$ is computed, which is given in (\ref{eqn_svd}).

\begin{equation}
\label{eqn_svd}
\mathbf{A} = \mathbf{U} \cdot \mathbf{\Sigma} \cdot \mathbf{V^*}
\end{equation}

\begin{figure}[!t]
\centering
\includegraphics[width=2.5in]{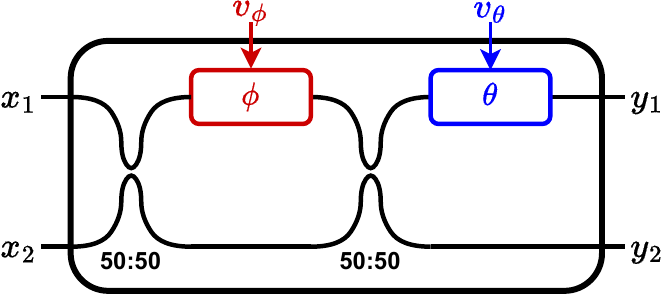}
\caption{Diagram of a MZI implementing a $2\times 2$ unitary transformation.}
\label{fig_simple_MZI}
\end{figure}

Note that $\mathbf{U}$ is a $M\times M$ unitary matrix, $\mathbf{\Sigma}$ is a $M\times N$ rectangular diagonal matrix with the singular values of $\mathbf{A}$ and $\mathbf{V^*}$ is the complex conjugate of the $N\times N$ unitary matrix $\mathbf{V}$. $\mathbf{\Sigma}$ can be implemented using optical attenuators for singular values less than $1$ and optical amplifiers for singular values greater than $1$ \cite{Shen2017}. Any unitary transformation such as $\mathbf{U}$ or $\mathbf{V^*}$ can then be implemented using a network of MZIs \cite{Reck1994}. Universal optical unitary multipliers necessitate the use of MZIs with two phase shifters \cite{Pai2020}, such as ones that consist of two 50:50 directional couplers and two tunable phase shifters arranged in an alternating order as shown in Fig. \ref{fig_simple_MZI}. Even when fabrication tolerances are neglected, the directional couplers have wavelength-dependent coupling ratios, resulting in a spectral response that is not constant. For the design wavelength $\lambda_0$ where the couplers provide an even power split, the output of an ideal MZI is given in (\ref{eqn_mzi}).

\begin{equation}
\label{eqn_mzi}
\begin{bmatrix}
y_1\\ 
y_2
\end{bmatrix} = \frac{1}{2}\begin{bmatrix}
e^{i\theta}(e^{i\phi}-1) & ie^{i\theta}(e^{i\phi}+1)\\ 
i(e^{i\phi}+1) & -(e^{i\phi}-1)
\end{bmatrix} \begin{bmatrix}
x_1\\ 
x_2
\end{bmatrix}
\end{equation}

Note that $\mathbf{x} = [x_1, x_2]^T$ are the complex fields corresponding to the MZI input signals, $\mathbf{y} = [y_1, y_2]^T$ are the complex fields corresponding to the MZI output signals, and $\phi$ and $\theta$ are the phase shifts introduced by the phase shifters as shown in Fig. \ref{fig_simple_MZI}. When using thermo-optic phase shifters for the MZI, the phase shifts can be expressed as a function of the applied voltage using:

\begin{equation}
\label{eqn_v2}
\begin{split}
\phi = \phi^{(2)} v_{\phi}^2 + \phi^{(0)} \\
\\
\theta = \theta^{(2)} v_{\theta}^2 + \theta^{(0)}
\end{split}
\end{equation}

For phase shift $\phi$, $v_{\phi}$ is the voltage applied to the heater, $\phi^{(2)}$ is the power to phase conversion ratio and $\phi^{(0)}$ is the phase shift when the voltage is set to 0 V. A similar notation was used for phase shift $\theta$.

It has been shown in \cite{Shen2017} that an arbitrary $2\times 2$ rotation matrix can be implemented by adjusting $\phi$ and $\theta$ using the corresponding heater voltages. It is necessary to have both phase shifters for independent control of amplitude and phase. Using MZIs in a network topology such as a triangular \cite{Reck1994} or a square mesh \cite{Clements2016}, it is possible to implement any arbitrary $N\times N$ unitary transformation such as $\mathbf{U}$ or $\mathbf{V^*}$.

\subsection{Experimental Setup for OMM}

\begin{figure*}[!t]
\centering\includegraphics[scale=0.75]{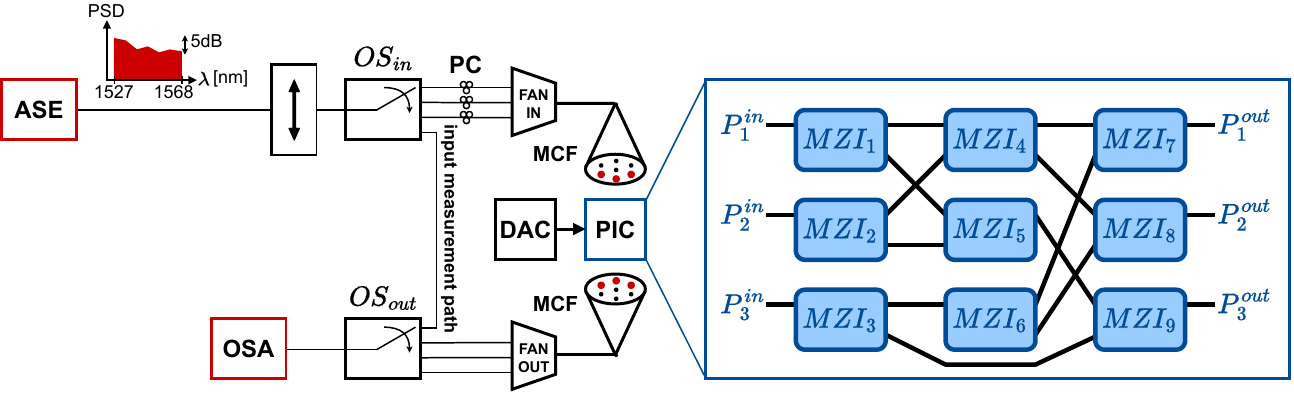}
\caption{Experimental measurement setup for the PIC. ASE: amplified spontaneous emission, OS: optical switch, PC: polarization control, MCF: multi-core fiber, OSA: optical spectrum analyzer, DAC: Digital-to-analog converter. The 3 red dots in the input and output MCFs indicate the cores that were used. Inset: MZI mesh configuration for OMM.}
\label{fig_exp_setup}
\end{figure*}

A portion of the silicon PIC described in \cite{Ding2016} was used for experimental evaluation of the models described in this work. This portion that was used for the $3\times 3$ OMM features 9 MZIs with a single titanium heater on one of the arms. Thus, a single heater voltage corresponding to $V_{\phi}$ in (\ref{eqn_v2}) is applied to each of the MZIs and it is not possible to modify the amplitude and the phase of an MZI output independently as $\theta$ is constant (no tunable phase shifter). Therefore, only the optical signal powers have been considered, which means that only a subset of all unitary transformations can be realized with this chip.

The experimental measurement setup for the PIC is shown in Fig. \ref{fig_exp_setup}. A 5-THz wide amplified spontaneous emission (ASE) source with a flattened power spectral density (PSD, variations $\leq$ 5~dB overall) was used as input probe to characterize the PIC. A $1\times 4$ optical switch ($OS_{in}$) allows to sequentially probe the three inputs of the chip which are accessible through vertical grating couplers. A bypass path - not coupled through the PIC - was used to record the input PSD. The grating couplers' arrangement on the PIC allows for probing it with a 7-core multi-core fiber (MCF), thus the outputs of the optical switch are connected to the three inputs of a 3-D inscribed fan-in device coupling from single-mode inputs to a MCF output~\cite{Ding2016}. In this work, only 3 out of the 7 cores were used. The grating couplers have an insertion loss of approx. 2 dB and are polarization sensitive, thus the broadband input is polarized and the 3 inputs used are individually polarization-aligned with polarization controllers (PC) to maximize transmission through the chip. The optical power at the output of the input MCF was kept to approx 12 dBm such that the optical power into the silicon optical waveguides does not give rise to nonlinear effects and the operation of the chip can be considered linear. The heaters on the PIC were wire-bonded to a printed circuit board and digital-to-analog converters (DACs) were used to independently control the voltages applied to each phase shifter. \ul{The chip was mounted on a thermally controlled stage for thermal stability.} At the output of the chip, a second set of grating couplers - with identical geometry - was used to out-couple the light onto a second MCF followed by a fan-out device whose 3 relevant outputs were connected to a $4\times 1$ optical switch ($OS_{out}$). $OS_{out}$ was then connected to an optical spectrum analyzer (OSA) which was used to capture the spectral responses of the matrix weights. The overall insertion loss for the 9 paths were between 9 and 11 dB - including the grating couplers - for voltages tuned at maximum transmission for the path under test. \ul{For voltage control, we used Qontrol Q8iv drivers with a voltage control resolution below 0.2 mV and for output monitoring, we used the Anritsu MS9740A OSA with a level precision of approximately 0.2 dB.} Finally, our chip uses multi-mode interferometer couplers (MMIs) instead of directional couplers to flatten the spectral response. Additional details on the design and fabrication of the photonic chip used for the measurements can be found in \cite{Ding2016}.

The measurements were carried out by fixing a set of voltages applied to the 9 MZIs under test and sequentially measuring the optical PSD at the $j^{th}$ input $P^{in}_{j}$ (constant throughout the measurements), the output PSD at the $i^{th}$ output $P^{out}_{i}$ and using them to calculate the implemented matrix weight using $w_{i,j} = \frac{P^{out}_{i}}{P^{in}_{j}}$. The PSDs measured by the OSA  were downsampled to 100 wavelengths corresponding to the central wavelengths of the ITU DWDM grid for the C-band with 50 GHz spacing.  Remark that any additional MZI traversed by the light on a given path but not directly involved in the $3\times 3$ OMM under test were kept at maximum transmission with their driving voltage constant throughout the measurements to ensure that the measured signal powers are as high as possible.

In order to generate a dataset for evaluating the MZI models, matrix weights resulting from a variety of voltages must be measured. The applied voltages were varied from 0 to 2 V corresponding to one half-period of the MZIs in order to prevent many-to-one mappings. Each MZI voltage was first swept through the entire range individually with a step size of 0.1 V while the other heater voltages were kept constant, resulting in 189 measurement points each consisting of the 9 matrix weights. The constant voltage levels were determined using individual voltage sweeps such that the diagonal matrix weights are maximized. Such a dataset is especially useful for training simple analytical models and it will be referred to as $\mathcal{D}_{sweep} = \{\mathbf{v}_{9\times1}^{(l)},\mathbf{W}^{(l)}_{3\times3\times N_{\lambda}} | l=1, ... , 189\}$. Note that the voltages $\mathbf{v}^{(l)}$ are in V while the matrix weights $\mathbf{W}^{(l)}$ are in dB. $N_{\lambda}$ is the number of discrete wavelengths in the spectrum. Then, 5100 additional measurements were obtained by applying random sets of 9 voltages sampled from 9 independent uniform distributions from 0 to 2 V. These additional measurements are needed to train NN models and they are also useful in fine-tuning analytical models. The additional measurements were split into a training set $\mathcal{D}_{training}$ with 70\%, a validation set $\mathcal{D}_{validation}$ with 15\%, and finally a testing set $\mathcal{D}_{testing}$ with the remaining 15\% of the datapoints. $\mathcal{D}_{validation}$ was included in $\mathcal{D}_{training}$ when a validation set was not necessary for training a model. 

\ul{The experimental measurements for the entire dataset took approximately 42 hours, 31 hours for the training set and 11 hours for the validation and testing sets. The main bottleneck was the OSA scanning time as a high sensitivity was required to ensure a sufficiently high dynamic range for the weights.}

The entire measurement procedure was repeated 6 times using the same heater voltages, and the acquired weights over the 6 runs were averaged to further reduce measurement uncertainty. \ul{The standard deviation of the output power for a 50 GHz spectral band over the 6 measurements is 0.7 dB.} The future use of pigtailed PICs with in- and out-coupling fibers glued to the silicon chip is expected to remove the need for measurement averaging, accelerating the dataset collection and potentially further reducing the modeling error.

\section{Modeling of MZI Meshes}

A model describing the relation between the heater voltages $\mathbf{v}^{(l)}$ and the implemented weight matrix $\mathbf{W}^{(l)}$ is required in order to program the PIC for OMM. Offline training of a (forward) model refers to finding the optimal parameters through the use of experimental measurements so that $\mathbf{W}^{(l)}$ can be predicted given $\mathbf{v}^{(l)}$. The more practical inverse model that can predict the necessary voltages such that a given matrix $\mathbf{W}^{(l)}$ is implemented is more difficult to obtain for the case of MZI meshes due to many-to-one mappings i.e. many different voltages can be used to realize the same matrix multiplication. So, in order to program an MZI mesh without the inverse model, the output of the forward model can be set to the desired output using an optimization algorithm. As the NN models presented in this work are differentiable, they also enable the use of powerful gradient-based optimizers \cite{Wright2022}. \ul{However, due to the periodic response of the MZIs with the phase difference between the arms, there is a chance that a voltage that is higher than necessary is used when this method is employed. The gradient-based optimizer could include an additional term that penalizes higher voltages to the cost function, ensuring minimum power consumption.}

All PIC models were trained by minimizing the root-mean-square error (RMSE) between the weights predicted by the model $\mathbf{\widehat{W}}^{(l)}$ and the experimentally measured target weights $\mathbf{W}^{(l)}$ in dB, which is given in (\ref{eqn_rmse}) for a dataset with $L$ datapoints. $w^{(l)}_{i,j,k}$ refers to the matrix weight with row index $i$ and column index $j$ for spectral slice $k$ with center wavelength $\lambda_k$ for datapoint $l$ in dB.

\begin{equation}
\label{eqn_rmse}
RMSE = \sqrt{\frac{\sum_{i=1}^{3}\sum_{j=1}^{3}\sum_{k=1}^{N_{\lambda}}\sum_{l=1}^{L}\left(\widehat{w}^{(l)}_{i,j,k}-w^{(l)}_{i,j,k}\right)^2}{9 \cdot L \cdot N_{\lambda}}}
\end{equation}

\subsection{Single-Wavelength Modeling}

The optical power within a $50$ GHz band around a center wavelength $\lambda_c = 1550$ nm was integrated for both the inputs and the outputs of single-wavelength models. Alternatively, the total power integrated across the entire C-band can also be used, as was the case in \cite{Cem2022}.

\subsubsection{Analytical models}

Simple analytical expressions such as (\ref{eqn_mzi}) can be used together with (\ref{eqn_v2}) to express $\mathbf{W}$ as a function of $\mathbf{v}$. (\ref{eqn_model1}) describes the simple analytical model (SAM) that also includes the finite extinction ratio of the MZIs, $ER$, and the optical losses, $\alpha_{i,j}$, corresponding to each path connecting input $j$ to output $i$.
\begin{equation}
w_{i,j} = \alpha_{i,j} \prod_{m \in M_{i,j}} \frac{1}{4} \left| \frac{\sqrt{ER} - 1}{\sqrt{ER} + 1} \pm e^{i \phi_m} \right|^2,
\label{eqn_model1}
\end{equation}
where $\phi_m = \phi^{(0)}_m + \phi^{(2)}_m v_m^2$ as in (\ref{eqn_v2}). $M_{i,j}$ denotes the set of all MZIs on the path connecting input $j$ to output $i$. The $\pm$ sign is a $+$ when MZI $m$ is used in the cross state and a $-$ when it is used in the bar state. As an example, $M_{2,1}$ would include MZIs 1 (bar), 4 (cross) and 8 (bar) for the MZI mesh configuration shown in Fig. \ref{fig_exp_setup}. Note that the initial phase offset $\phi^{(0)}_m$ and the power to phase conversion ratio $\phi^{(2)}_m$ for MZI $m$ should be trained individually for each of the MZIs to properly account for fabrication imperfections. While this can also be done for $ER$, this only provided a negligible improvement to modeling accuracy for this particular PIC as the extinction ratios for the MZIs in the fabricated PIC are very similar. Therefore, the average value of $ER_{avg} = 30$ dB was used for all MZIs. When training SAM, $\mathcal{D}_{sweep}$ was used to fit $\phi^{(0)}_m$, $\phi^{(2)}_m$ and $ER$. Then, $\mathcal{D}_{training}$ was used to optimize $\alpha_{i,j}$. Both the initial training using $\mathcal{D}_{sweep}$ and the secondary optimization using $\mathcal{D}_{training}$ was performed using the unconstrained multivariable optimization function \textit{fminunc()} in MATLAB\textsuperscript{TM}, which makes use of the BFGS algorithm.

For the PIC under investigation, SAM is equivalent to having 3 MZIs connected in series for each of the matrix weights as the MZI mesh is organized in 3 layers of 3 MZIs. SAM can accurately predict $\mathbf{W}$ within $\mathcal{D}_{sweep}$, where only 1 voltage is varied at a time. However, the same cannot be said for other datapoints where multiple voltages are varied simultaneously such as the ones in $\mathcal{D}_{training}$. This is due to the fact that the heaters apply undesired phase shifts for neighboring MZIs through thermal crosstalk, which cannot be accounted for using SAM as each matrix weight only depends on the 3 voltages on the corresponding path. In order to have a simple analytical model that can account for thermal crosstalk, we introduce a second model SAM+XT, which uses the Thermal Eigenmode Decomposition method \cite{Milanizadeh2019,Perez2020}. The equation describing SAM+XT closely resembles the one for SAM given in (\ref{eqn_model1}), but it also attempts to correct for thermal crosstalk by including contributions from all voltages for the phase shifts without characterizing the crosstalk explicitly. This is done by introducing additional fitting parameters for $\phi_m$, resulting in the new expression given in (\ref{eqn_model2}).
\begin{equation}
\phi_m = \phi^{(0)}_{m} + \sum_{n=1}^{N_{MZI}} \phi^{(2)}_{m,n} v_n^2
\label{eqn_model2}
\end{equation}
Note that $N_{MZI}$ is the total number of MZIs in the entire PIC. When training SAM+XT, the optimal values of the parameters $\phi^{(0)}_m$, $\phi^{(2)}_{m}$ and $ER$ for SAM were used as an initial point. Note that the thermal crosstalk terms $\phi^{(2)}_{m,n} \; (m\neq n)$ were set to $0$ while the diagonal entries $\phi^{(2)}_{m,m}$ were set to $\phi^{(2)}_{m}$. Then, $D_{training}$ was used to optimize $\alpha_{i,j}$ and $\phi^{(2)}_{m,n}$ using the BFGS algorithm. SAM+XT is an extension to SAM and can also be used for cases where thermal crosstalk is not severe, as setting the thermal crosstalk terms $\phi^{(2)}_{m,n} = 0$ makes it equivalent to SAM.
 
\subsubsection{Data-driven models}

\begin{figure}[!t]
\centering
\includegraphics[scale=0.6]{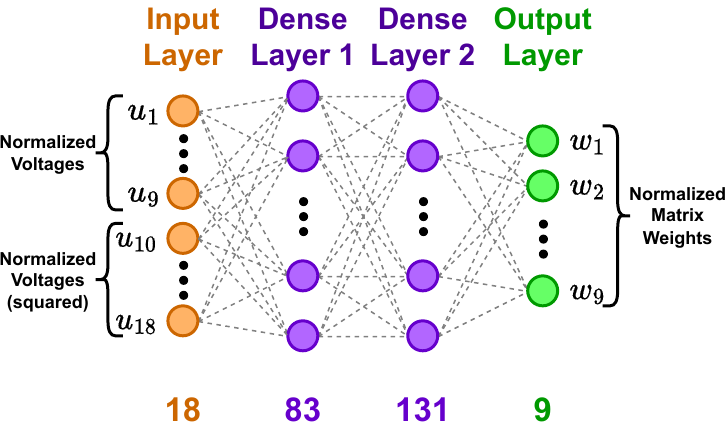}
\caption{Architecture of the single-wavelength NN model of the PIC. Numbers at the bottom indicate the number of nodes in the corresponding layer.}
\label{fig_model3}
\end{figure}

Alternatively, the MZI mesh can be modeled using a NN without relying on physics-based analytical expressions. Nevertheless, some physical intuition from (\ref{eqn_v2}) was included in the model by including the squares of the heater voltages alongside the heater voltages themselves at the input, increasing the total number of inputs to 18. This new input is referred to as $\mathbf{u} = [\mathbf{v}, \mathbf{v}^2]^T$. The architecture for the single-wavelength NN model (NN-SW) is shown in Fig. \ref{fig_model3}.

Both the inputs $\mathbf{u}$ and the outputs $\mathbf{W}$ were rescaled between $[-1,+1]$ through min-max normalization for all data-driven models. The first and the last 9 elements of $\mathbf{u}$ were normalized individually as their ranges before normalization are different. All outputs were denormalized by applying the inverse linear transformation used for normalization when evaluating the models e.g. when calculating the RMSE. The outputs were also flattened into a column vector $\mathbf{w}^{(l)}_{9\times 1}$ with elements $w^{(l)}_{p}$ where $p$ is the 1-D index. The PyTorch ML framework was used for training and all data-driven models were trained using the L-BFGS optimizer. The hyperbolic tangent activation function was used for all hidden layers. The activation function, the number of hidden layers as well as the number of nodes used at each hidden layer were determined through hyperparameter optimization using the tree-structured Parzen estimator implemented within the Optuna framework \cite{Akiba2019}. $\mathcal{D}_{training}$ was used for training the model while $\mathcal{D}_{validation}$ was used for hyperparameter optimization and for implementing a stopping criterion for training, where training was stopped when the performance on $\mathcal{D}_{validation}$ did not improve by more than 0.001 dB for 50 consecutive epochs.

\subsection{Multiple-Wavelength Modeling}

\begin{figure}[!t]
\centering
\includegraphics[scale=0.15]{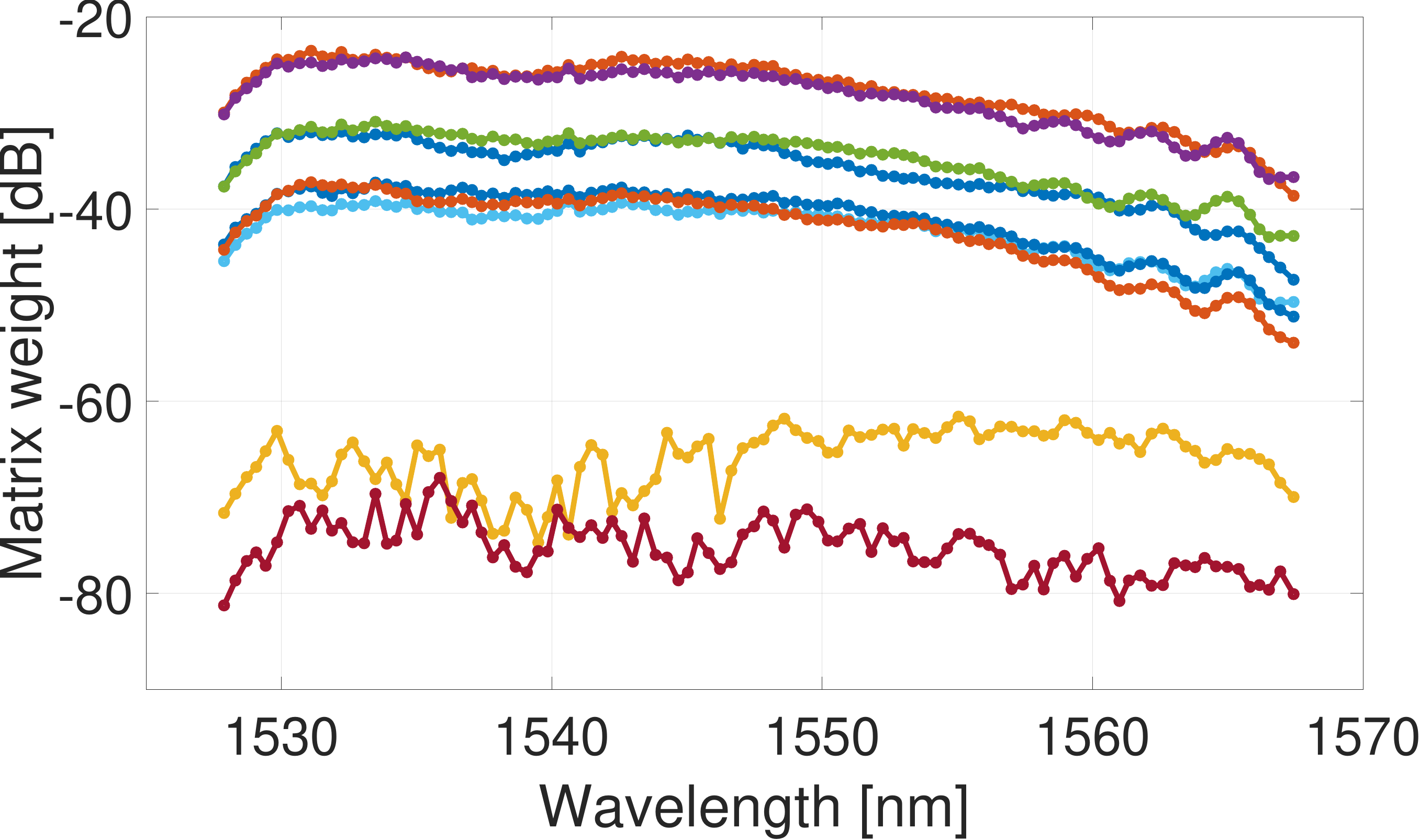}
\caption{Sample raw spectral measurement results for a single set of voltages, each of the 9 curves represents a separate matrix weight where each dot corresponds to 1 of the 100 center wavelengths.}
\label{fig_samp_meas}
\end{figure}

\begin{figure}[!t]
\centering
\includegraphics[scale=0.15]{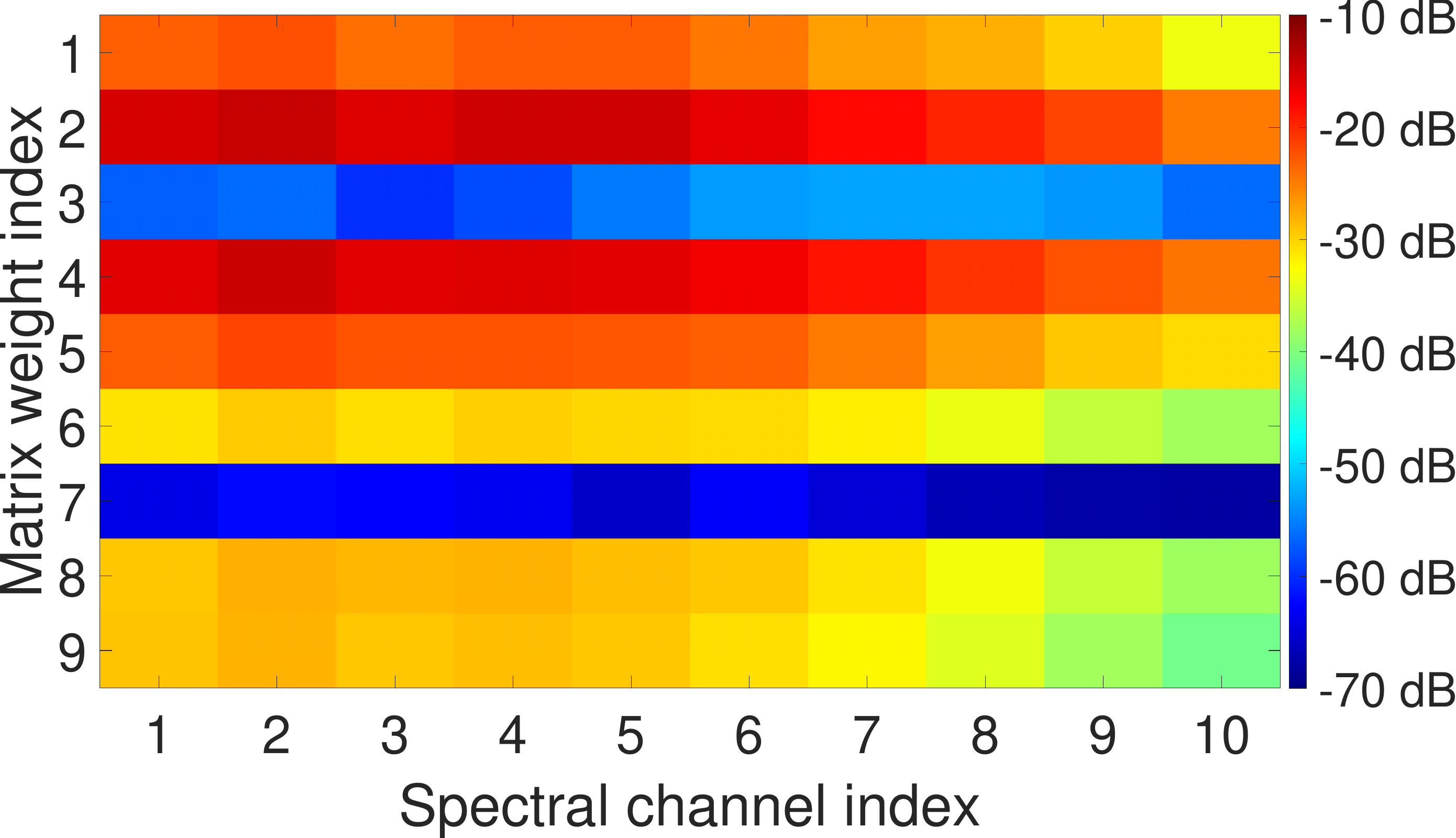}
\caption{Sample spectral measurement results for a single set of voltages, downsampled to 10 spectral channels by integration and reorganized into a 2D matrix form.}
\label{fig_samp_meas_img}
\end{figure}

While MZI meshes have mostly been used for coherent single-wavelength OMM, they have recently been proposed for applications making use of wavelength division multiplexing as well, which has the potential to provide extended parallelization capabilities~\cite{Totovic2022}. In this section, four different NN-based models for multiple-wavelength modeling will be described. The number of wavelengths $N_{\lambda}$ was set to 10 by integrating the power across groups of 10 spectral bands (post-processing), as it is challenging to independently model the 100 frequency channels for some of the models under investigation. After downsampling, the measured weights for a single $\mathbf{v}^{(l)}$ were rearranged into a $9\times 10$ 2-D form, where each column corresponds to a spectral channel and each row corresponds to a single matrix weight. The processed 2D-output $\widetilde{\mathbf{W}}^{(l)}$ consists of elements $\widetilde{w}^{(l)}_{p,k} = w^{(l)}_{p}(\lambda_k)$ and its values for the measured responses in Fig. \ref{fig_samp_meas} are shown in Fig. \ref{fig_samp_meas_img}.

Observing the curves in Fig. \ref{fig_samp_meas}, the spectral responses of the 9 weights appear to be similar across the C-band apart from vertical shifts. Inspired by this, the NN model with wavelength rescaling (NN-$\lambda$R) attempts to use a NN to predict $\mathbf{w}^{(l)}(\lambda_c)$ and rescale the weights differently for each wavelength using a scaling factor $b(\lambda_k)$ with $b(\lambda_c)=0$ as shown in (\ref{eqn_modelb}).

\begin{equation}
\label{eqn_modelb}
w^{(l)}_p(\lambda_k) = w^{(l)}_p(\lambda_c) \cdot b(\lambda_k)
\end{equation}

\begin{figure}[!t]
\centering
\includegraphics[scale=0.4]{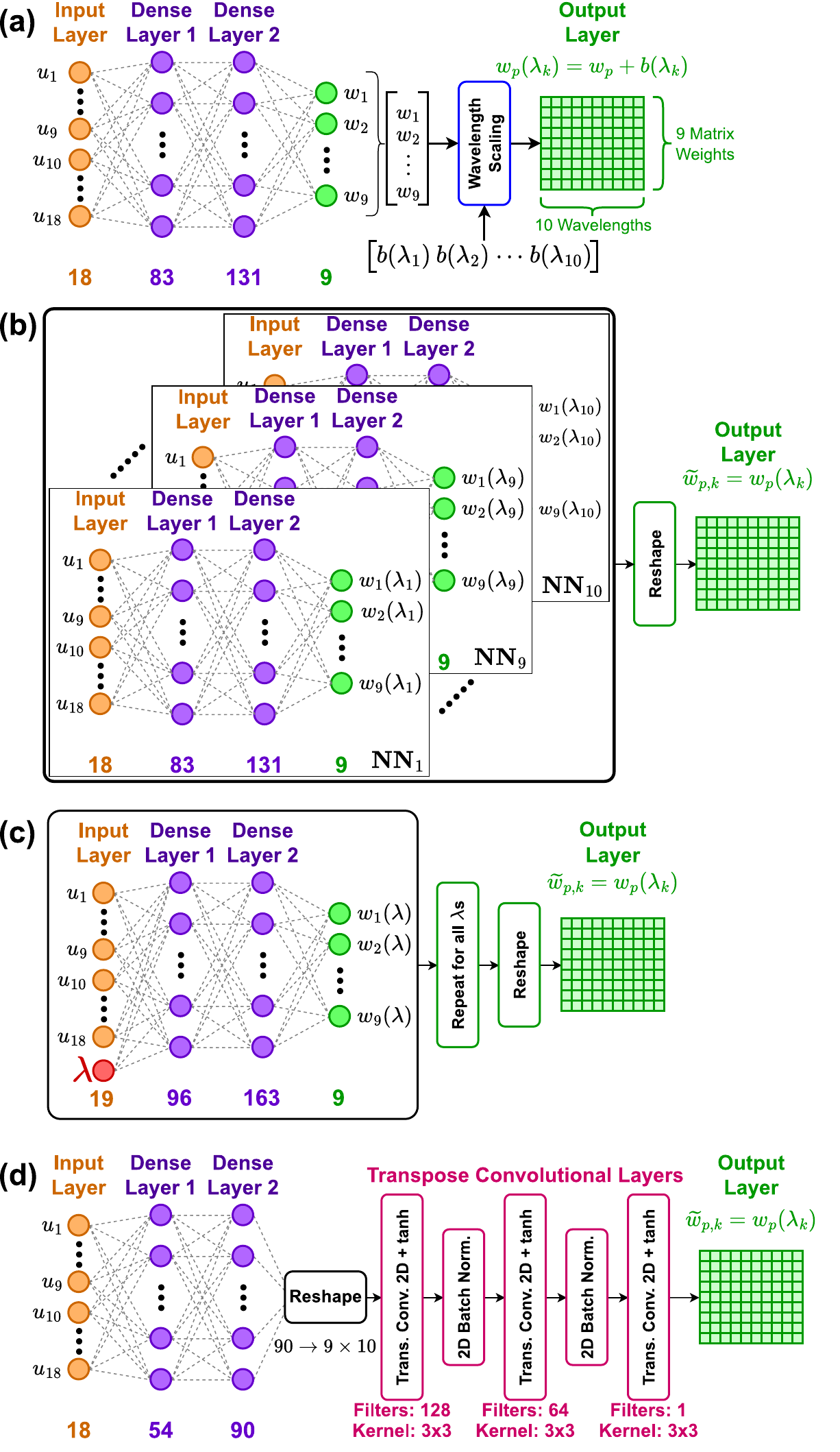}
\caption{Architectures of the multiple-wavelength models of the PIC. (a) NN-$\lambda$R, (b) NN-$\lambda$S, (c) NN-$\lambda$G, (d) TCNN.}
\label{fig_model_lambda}
\end{figure}

Note that NN-SW was trained for a 50 GHz spectral interval while NN-$\lambda$R was trained for a 500 GHz interval. Fig. \ref{fig_model_lambda}(a) illustrates the architecture for NN-$\lambda$R. $D_{training}$ was first used to train the NN for the spectral slice closest to the center of the C-band, which is $\lambda_6 = 1550$ nm. Then, the optimal values for $b(\lambda_k)$ were found after training the NN using the BFGS algorithm such that the validation error is minimized for each $k$. While this approach can correct for wavelength-dependent losses in the chip, it cannot model the chip accurately if the nature of the relation between $\mathbf{v}$ and $\mathbf{W}$ is wavelength dependent.

The wavelength-specific NN model (NN-$\lambda$S) consists of 10 individual NNs with the same structure as NN-SW, each trained using the weights at a different part of the spectrum $\mathbf{w}(\lambda_k)$. This allows the model to learn any differences for the spectral responses of the individual matrix weights. The optimal hyperparameters are found to be very similar for all 10 NNs and the same hyperparameters were used for all of them as shown in Fig. \ref{fig_model_lambda}(b). Each NN was trained individually in similar fashion to NN-SW and the resulting $\mathbf{w}^{(l)}(\lambda_k)$ were combined to obtain $\mathbf{W}^{(l)}$ during inference, which may not be a practical approach for cases where $N_{\lambda}$ is much higher than 10.

For the general wavelength NN model (NN-$\lambda$G), the wavelength $\lambda$ is added as an additional input after normalization, resulting in the architecture shown in Fig. \ref{fig_model_lambda}(c). This allows the model to learn the spectral behavior of the PIC simultaneously for all matrix weights and removes the need to train 10 individual NNs. However, the datasets must be modified slightly: each datapoint $l$ is divided up into 10 datapoints with the same $\mathbf{v}$ but different $\lambda$s, resulting in different outputs $\mathbf{w}(\lambda_k)$. Furthermore, the optimal number of nodes for the hidden layers are slightly higher compared to the ones used in earlier NN models, as indicated in Fig. \ref{fig_model_lambda}. As a result, the training procedure is still time-consuming mainly due to the fact that the training set is unnecessarily large, which is an issue that can be resolved by taking advantage of its highly correlated nature.

The correlation is indeed exploited in a transposed convolutional neural network (TCNN) model \cite{Zeiler2010} that utilizes transposed convolutional layers to take advantage of the fact that the matrix weights for adjacent spectral channels are correlated. Fig. \ref{fig_model_lambda}(d) shows the architecture found after hyperparameter optimization. Compared to the previous models, training TCNN neither requires the training of multiple NNs like for NN-$\lambda$S, nor does it necessitate an increase in the training set size like for NN-$\lambda$G. Therefore, TCNN scales well for higher values of $N_{\lambda}$ in the sense that the computational resources required for training does not increase dramatically. To demonstrate this, the TCNN model was also evaluated for the case without further downsampling the spectral responses i.e. $N_{\lambda} = 100$, also presented in \cite{CemIPC2022}. This model will be referred to as TCNN-100$\lambda$. Compared to the 10-wavelength TCNN model, the number of nodes in Dense Layer 1 is increased from 54 to 128 and the number of nodes in Dense Layer 2 is increased from 90 to 900 to obtain a $9\times N_{\lambda}$ output.

\section{Experimental Results and Discussion}

After training the models using the learning procedures described in the previous chapter, the models were evaluated experimentally using $\mathcal{D}_{testing}$. \ul{Training NN-SW takes less than 5 minutes on a laptop computer. In contrast, training SAM takes less than 10 seconds and training SAM+XT takes less than 30 seconds on the same computer. Once trained, 13,546 multiply and accumulate operations are required for inference i.e. to calculate the implemented weights for a given input voltage vector $\mathbf{u}$, which takes less than a microsecond on a standard computing unit. However, operation is in the kHz range for setting the heater voltages to new values due to the digital-to-analog converters and the slow thermal response of the heaters}

\subsection{Single-Wavelength Modeling}

\begin{figure}[!t]
\centering
\includegraphics[scale=0.15]{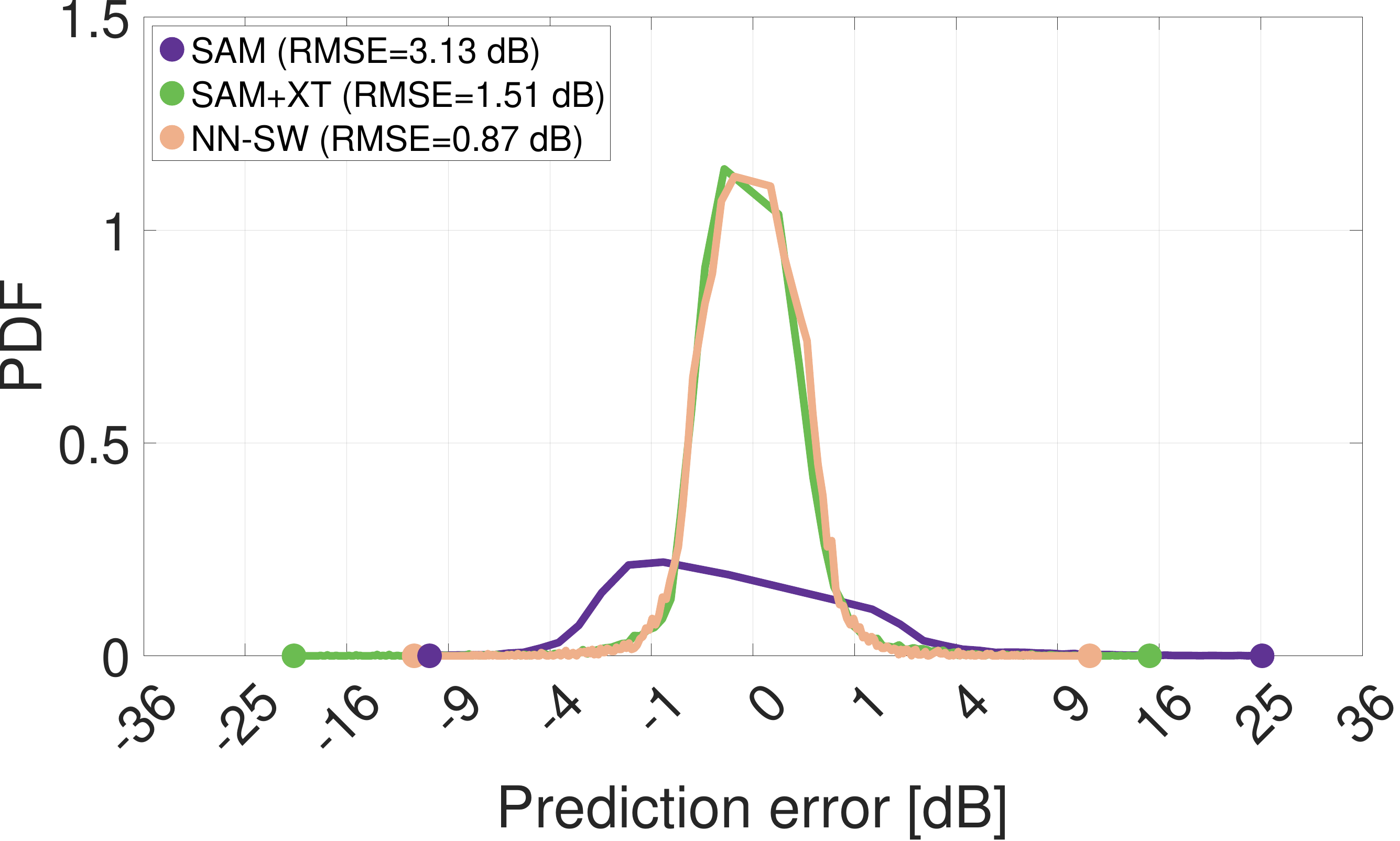}
\caption{Probability density functions obtained by normalizing the error histograms for the single-wavelength models. The plot has undergone a nonlinear transformation in the horizontal direction to better show the upper and lower limits for the errors. The large dots indicate the minimum and maximum errors for each curve.}
\label{fig_pdf_123}
\end{figure}

\begin{figure}[!t]
\centering
\includegraphics[scale=0.15]{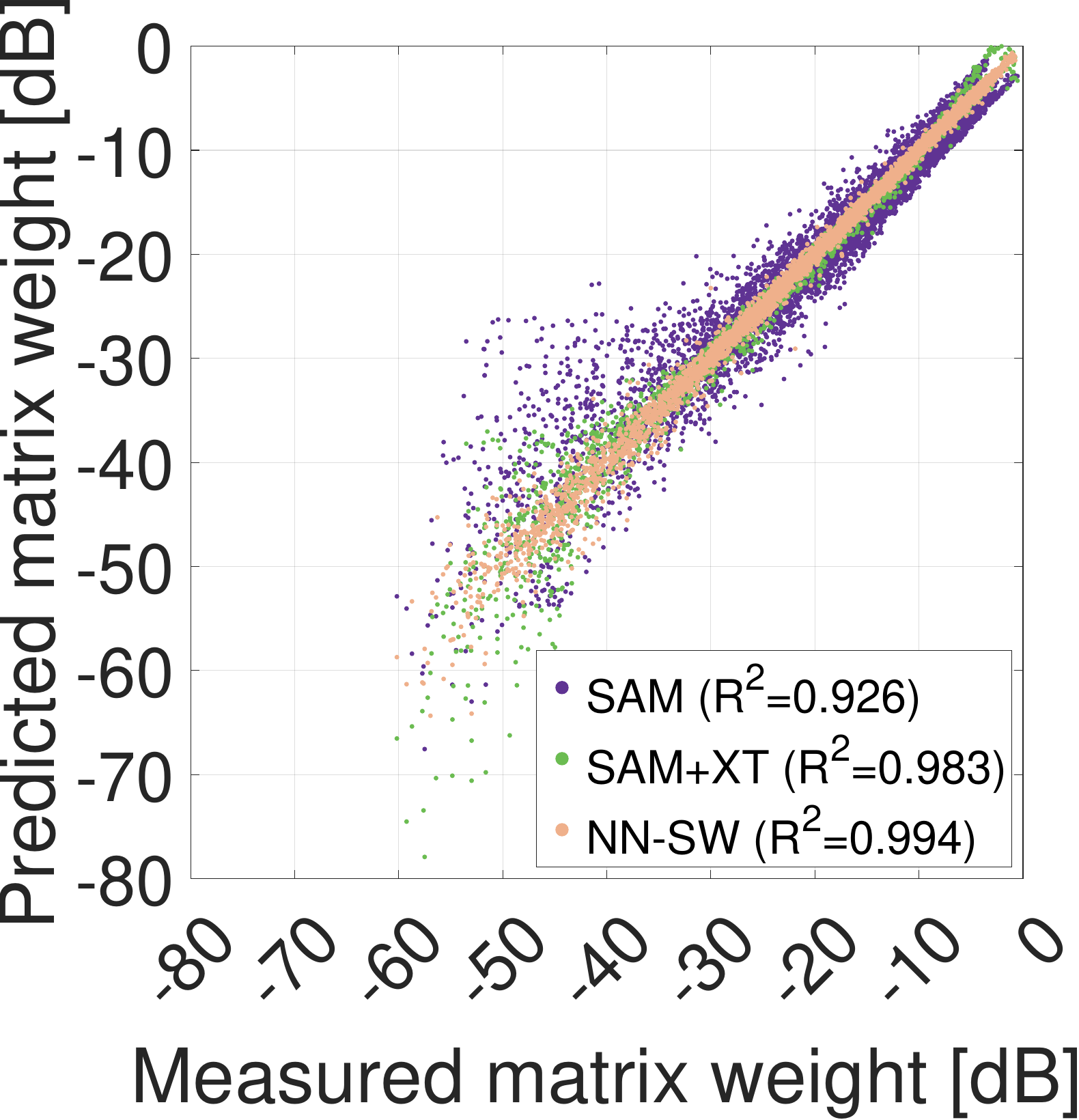}
\caption{Scatter plot of the predicted and measured matrix weights for the single-wavelength models. The coefficients of determination ($R^2$) for the models are provided in the legend.}
\label{fig_corr_curve_123}
\end{figure}

For the single-wavelength models, the RMSEs between the measured and the predicted matrix weights in $\mathcal{D}_{testing}$ were found to be 3.13 dB, 1.51 dB, and 0.87 dB  for SAM, SAM+XT, and NN-SW, respectively. Fig. \ref{fig_pdf_123} shows the probability distribution functions (PDFs) of the errors between predicted and measured weights. The addition of thermal crosstalk terms improves the simple analytical model, as SAM+XT is more accurate than SAM. However, similar to the results obtained in \cite{Cem2022}, the data-driven NN-SW outperforms the simple analytical models both in terms of maximum absolute error and the RMSE. There is a degradation in RMSE from 0.53 dB to 0.87 dB for NN-SW compared to the results presented in \cite{Cem2022}. This is due to the fact that a much wider spectral band has been used for modeling in \cite{Cem2022}, as integrating over a larger part of the spectrum reduces the relative impact of measurement uncertainty. The labels and predictions for all weights in $\mathcal{D}_{testing}$ are shown in Fig. \ref{fig_corr_curve_123}. All 3 models perform relatively well for weights close to 0 dB and the modeling performance decreases for the lower weights especially for SAM and SAM+XT as lower values result in more uncertainty. 

\subsection{Multiple-Wavelength Modeling}

\begin{figure}[!t]
\centering
\includegraphics[scale=0.15]{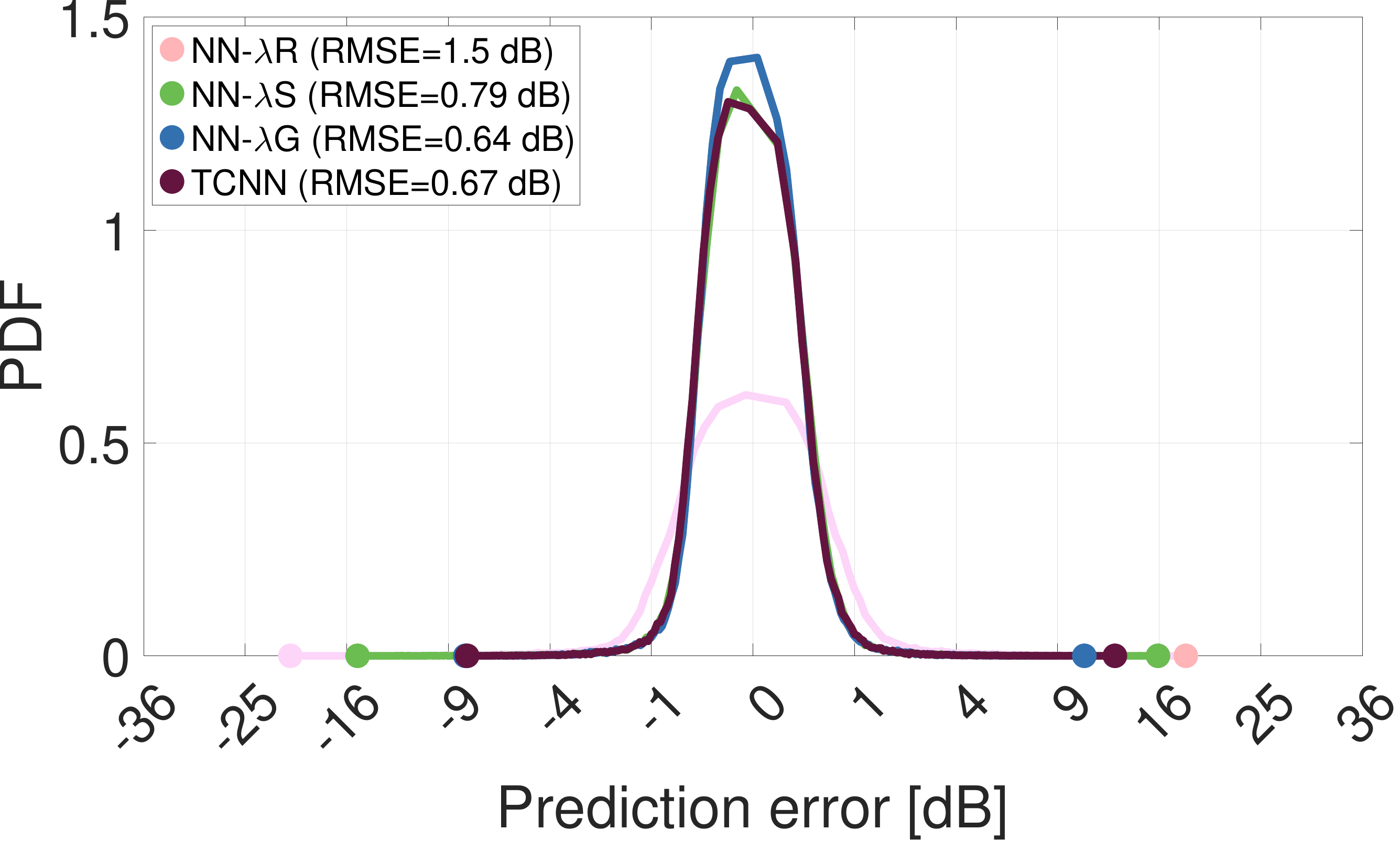}
\caption{Probability density functions obtained by normalizing the error histograms for the multiple-wavelength models.}
\label{fig_pdf_3bcde}
\end{figure}

The RMSEs between the measured and the predicted matrix weights in $\mathcal{D}_{testing}$ with $N_{\lambda} = 10$ were found to be 1.42 dB, 0.79 dB, 0.64 dB, and 0.67 dB  for NN-$\lambda$R, NN-$\lambda$S, NN-$\lambda$G, and TCNN, respectively. Fig. \ref{fig_pdf_3bcde} shows the PDFs of the errors between predicted and measured weights. NN-$\lambda$R performs the worst out of the 4, which suggests that using a constant scaling factor to model the response at each wavelength is not sufficient to obtain a good model. NN-$\lambda$S is able to achieve a considerably lower RMSE by training for each wavelength individually. However, the maximum absolute prediction error for NN-$\lambda$S is much higher than the ones obtained with NN-$\lambda$G and TCNN, which is the main reason why their corresponding RMSEs are lower than that of NN-$\lambda$S. Both models include data for different wavelengths simultaneously during the training of the NNs, which is not the case for NN-$\lambda$S.

\begin{figure}[!t]
\centering
\includegraphics[scale=0.15]{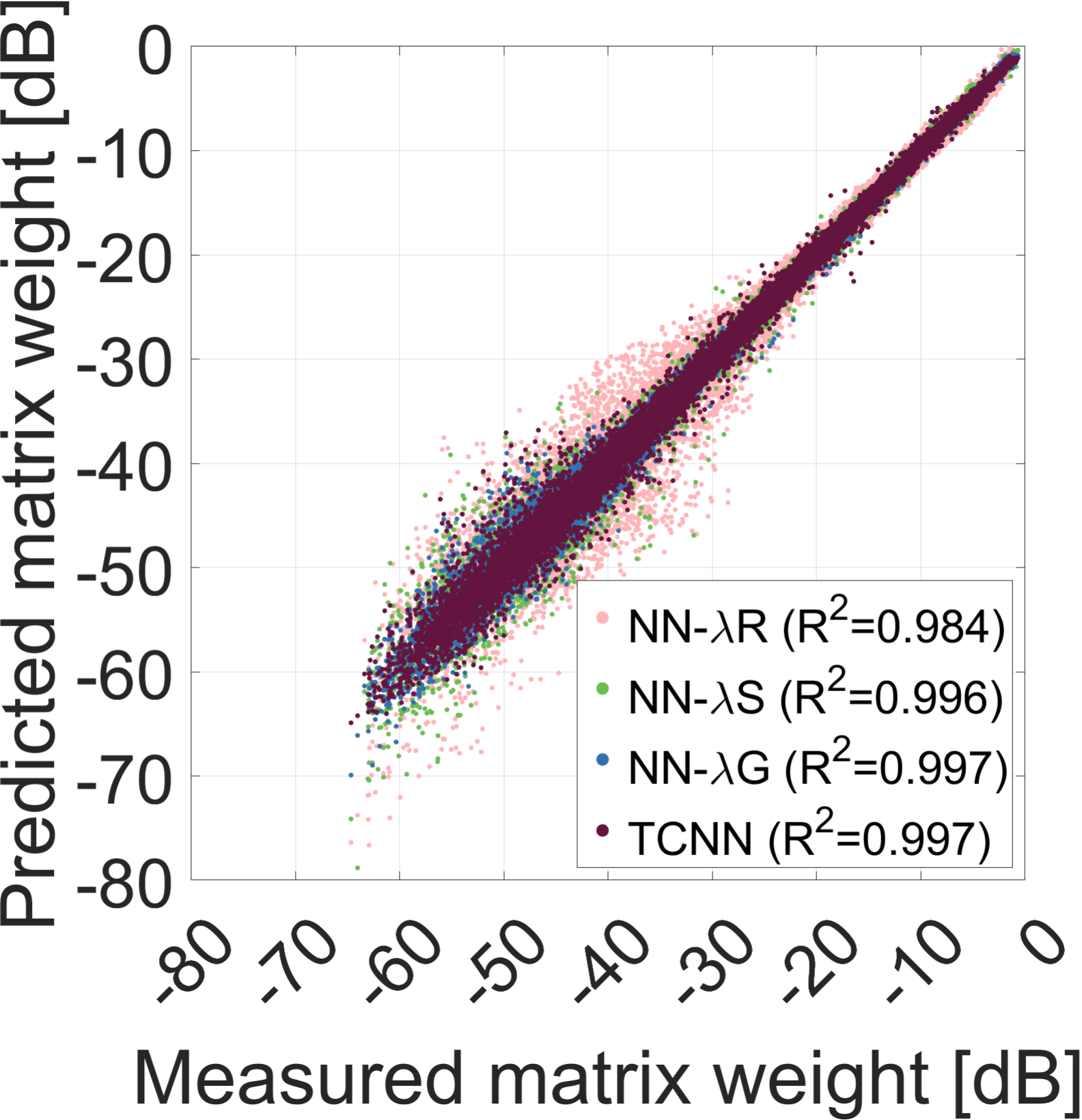}
\caption{Scatter plot of the predicted and measured matrix weights for the multiple-wavelength models. The coefficients of determination ($R^2$) for the models are provided in the legend.}
\label{fig_corr_curve_3bcde}
\end{figure}

Fig. \ref{fig_corr_curve_3bcde} shows the scatter plots for the predicted and measured weights for the four multiple-wavelength models. All four models perform exceptionally well for voltages resulting in weights close to $0$ dB and their performances degrade for low matrix weights.

\begin{figure}[!t]
\centering
\includegraphics[scale=0.15]{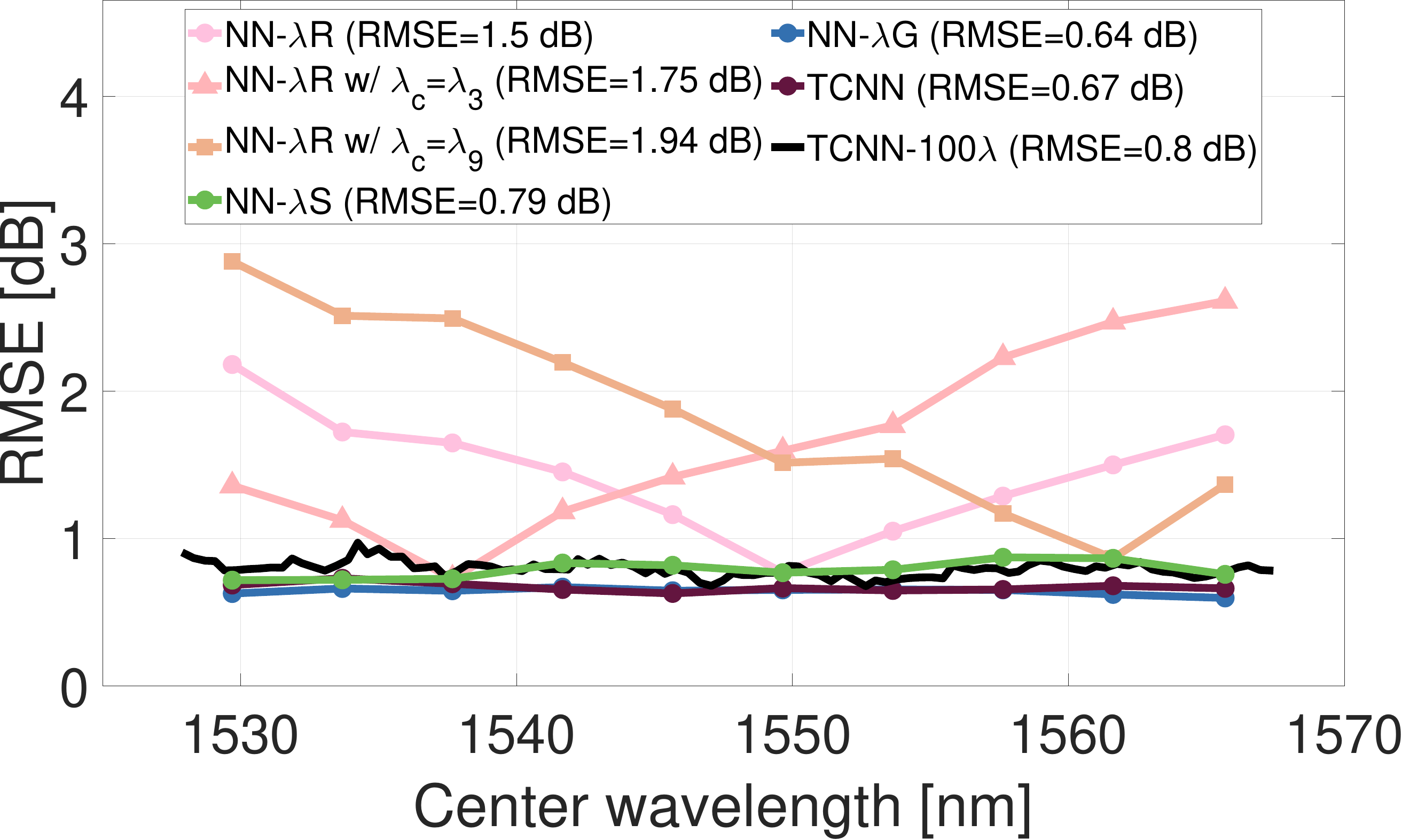}
\caption{Testing RMSEs obtained for spectral intervals with different center wavelengths for the multiple-wavelength models.}
\label{fig_rmse_lambda}
\end{figure}

Fig. \ref{fig_rmse_lambda} shows the performance of the multiple-wavelength models in terms of the testing RMSEs individually for the modeled wavelengths. While the RMSEs are close to flat for the other models, the performance depends heavily on the wavelength for NN-$\lambda$R. The testing RMSE for NN-$\lambda$R is low for the wavelength ($\lambda_6$) used to train the NN part of the model and gets higher as the model is used to predict the matrix weights at wavelengths that are further away from $\lambda_6$. Note that NN-$\lambda$R is identical to NN-$\lambda$S at $\lambda_6$. When another center wavelength is used to train the NN part, such as $\lambda_3$ or $\lambda_9$, the minimum RMSE is observed for the new center wavelength, but the V-shaped nature of the RMSE-$\lambda$ curve remains the same, resulting in higher overall RMSE when center wavelengths further away from $\lambda_6$ are used. The results show that there is more to multiple-wavelength modeling than rescaling the results obtained by a single-wavelength model, which has led to the use of the more complex architectures. Refer to Appendix A for further details on why this is the case.

\begin{figure}[!t]
\centering
\includegraphics[scale=0.15]{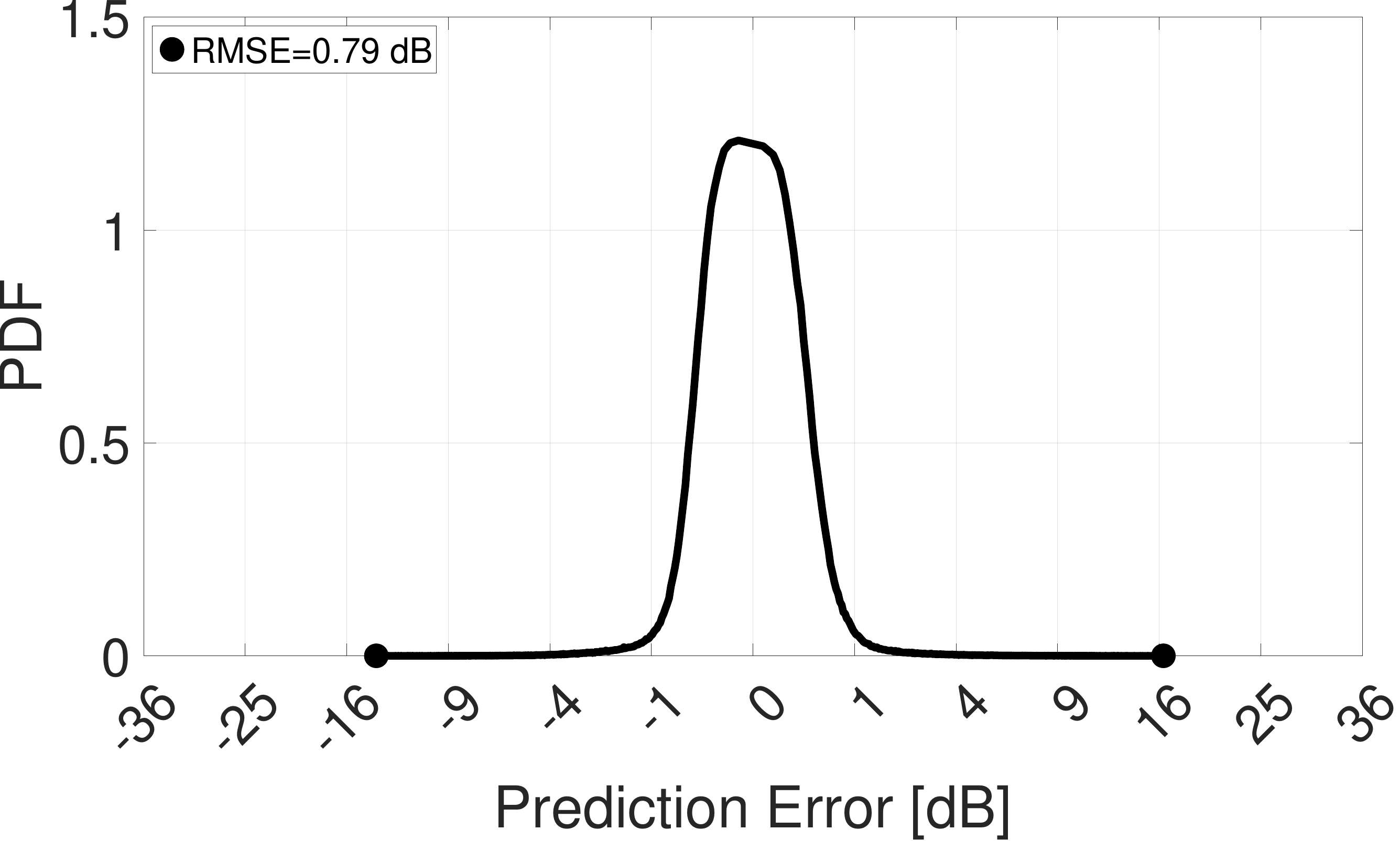}
\caption{Probability density function obtained by normalizing the error histograms for the TCNN-100$\lambda$ model}
\label{fig_pdf_TCNN}
\end{figure}

\begin{figure}[!t]
\centering
\includegraphics[scale=0.15]{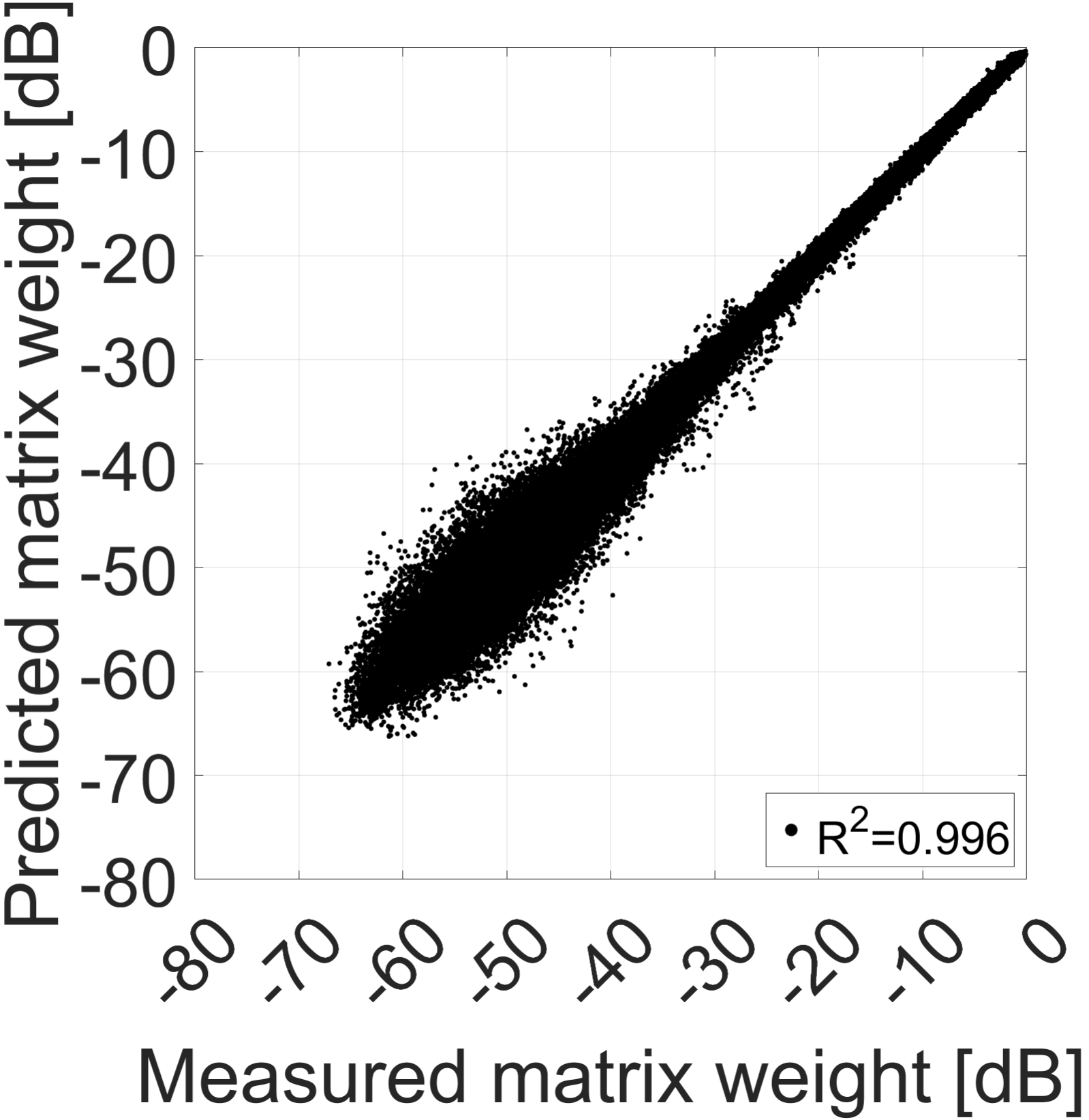}
\caption{Scatter plot of the predicted and measured matrix weights for the TCNN-100$\lambda$ model. The coefficient of determination ($R^2$) for the model is provided in the legend.}
\label{fig_corr_curve_TCNN}
\end{figure}

While NN-$\lambda$G and TCNN perform equally well for $N_{\lambda} = 10$, TCNN-100$\lambda$ is the practical choice for $N_{\lambda} = 100$. The error histogram for TCNN-100$\lambda$ is shown in Fig. \ref{fig_pdf_TCNN}. The RMSE was found to be 0.79 dB despite the higher impact of noise resulting from the narrower integration width. Furthermore, the model retains its impressive performance for high weights and only struggles to model the low weights (e.g. $w^{(l)}_{i,j,k} < -30$ dB) as displayed in the scatter plot in Fig. \ref{fig_corr_curve_TCNN}.

\subsection{Impact of Training Set Size on Modeling Performance}

\begin{figure}[!t]
\centering
\includegraphics[scale=0.15]{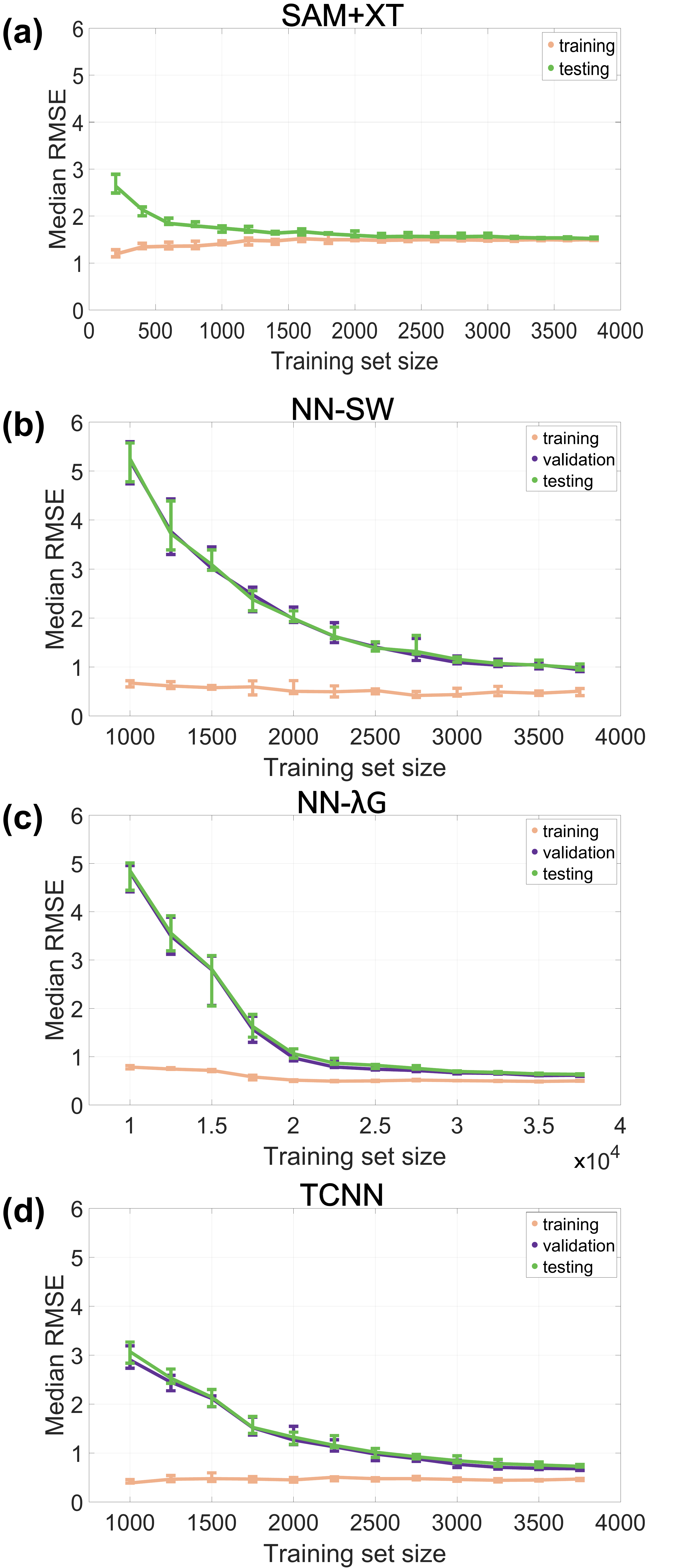}
\caption{Modeling performance in terms of RMSE on the training, validation, and testing sets as a function of the number of datapoints used during training for the \ul{(a) SAM+XT}, (b) NN-SW, (c) NN-$\lambda$G, and (d) TCNN models. Error bars show the $25^{th}$ and $75^{th}$ percentiles and the line curve shows the median RMSE for 10 different random seeds. Note that the dataset size was artificially increased by a factor of 10 as a result of defining each spectral measurement as a separate datapoint for NN-$\lambda$G.}
\label{fig_Ltrain}
\end{figure}

\ul{A certain number of training datapoints is required for any model to reach its peak performance. We investigated the relation between the training set size and modeling performance by training both the analytical and data-driven models using fewer measurements. 10 different models were trained each with a different random seed for all training set sizes under investigation and the datapoints were chosen randomly from all available measurements. For the NN models, the random seed has an additional impact on the performance due to the randomness in network initialization, separation of training data into mini-batches, and initialization of the Hessian matrix for the L-BFGS optimizer.} Note that the splitting of the dataset was not changed depending on the random seed. In order to demonstrate that a sufficient number of measurements have been performed, the training procedure was repeated for 10 different random seeds. The median RMSEs along with the $25^{th}$ and $75^{th}$ percentiles for training with different random seeds are shown in Fig. \ref{fig_Ltrain}(a), (b), (c), and (d) for SAM+XT, NN-SW, NN-$\lambda$G and TCNN, respectively. 
 
\ul{For SAM+XT, using 1000 datapoints from $\mathcal{D}_{training}$ for training in addition to the initial training of the phase parameters $\phi^{(0)}$ and $\phi^{(2)}$ using $\mathcal{D}_{sweep}$ is sufficient to achieve RMSEs within 0.2 dB of the reported performances.} NN-SW achieves close to peak performance after training for 3250 datapoints. While the effect of using different random seeds is high for lower training set sizes, the differences between the performances obtained with different seeds gradually diminish as the training set size is increased to around 3000. Similar observations can be made for NN-$\lambda$G after accounting for the fact that each datapoint is replaced by 10 datapoints for training, each corresponding to a different wavelength. Finally, TCNN performs significantly better than the other ML models when trained with fewer measurements. However, its performance converges to a similar level as the number of measurements reaches 3250.

The results indicate that performing around 3500 measurements is sufficient to train data-driven models for the PIC at hand. Obtaining additional datapoints does not provide an improvement in performance outside error bars for any of the models. If fewer datapoints are available, TCNN is the most promising model among the ones that have been analyzed. Nevertheless, simple analytical models achieve lower errors than the best performing data-driven model when a highly limited dataset with less than 1500 datapoints is available for training.

\subsection{Relating Modeling Error to Performance on a ML Task}

\begin{figure}[!t]
\centering
\includegraphics[scale=0.15]{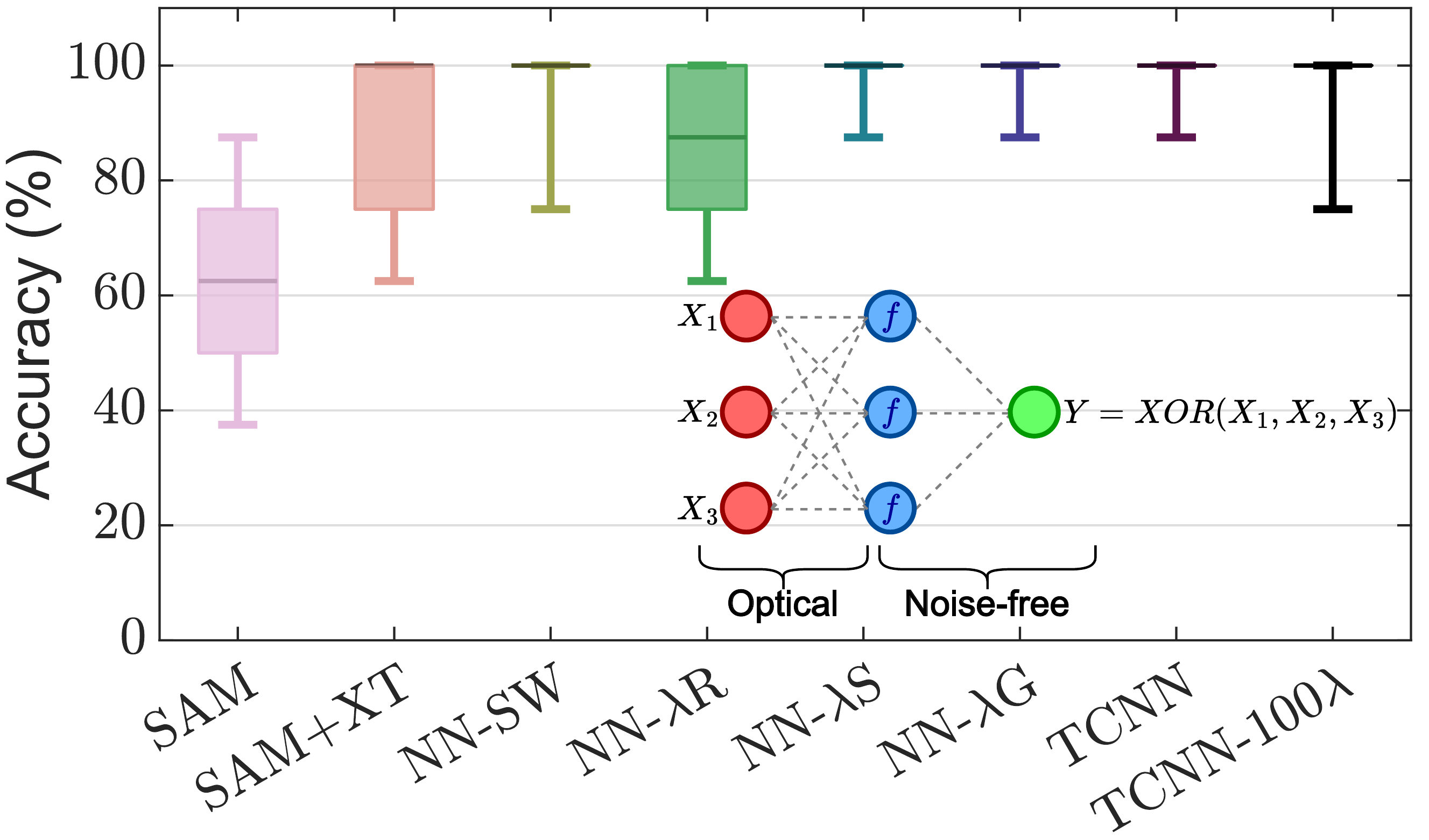}
\caption{Classification accuracy on the 3-input XOR task with noise added on the linear layer weights according to the prediction error distributions for the models. Boxes show the $25^{th}$ and $75^{th}$ percentiles while the whiskers show the $10^{th}$ and $90^{th}$ percentiles for 2000 noise realizations. The dark horizontal lines indicate the median accuracies. Inset: architecture of the NN.}
\label{fig_toy_acc}
\end{figure}

In order to quantify the practical advantage of improving modeling accuracy for relating the applied voltages to the obtained weights for OMM, the various models that were analyzed so far should be used to program a PIC for OMM to implement the linear layer of an optical NN. For the specific chip under investigation, a single-hidden-layer NN with 3 nodes in the input layer and 3 nodes in the hidden layer was considered. The inset in Fig. \ref{fig_toy_acc} shows the NN architecture. Note that the hyperbolic tangent activation functions were not shown explicitly. The NN was trained using PyTorch for the 3-bit XOR task described in \cite{Williamson2019}, for which it has an accuracy of 100\%. For this simple task at hand, the desired output of the NN is 1 only when exactly one of the inputs is equal to 1, and 0 otherwise.

We numerically emulated the case where the MZI mesh was used to implement the $3 \times 3$ linear layer optically. The impact of modeling error for the linear layer was introduced as noise to the weights, uniformly sampled from the testing error distributions for each model. The nonlinear activation functions and the $3\times 1$ linear output layer were kept noise-free so that the noise only emulates the the prediction errors between the physical chip and the models used for programming. The classification accuracies for the noisy NNs corresponding to the different models are shown in Fig. \ref{fig_toy_acc}. Boxes show the $25^{th}$ and $75^{th}$ percentiles while the whiskers show the $10^{th}$ and $90^{th}$ percentiles for 2000 noise realizations. The dark horizontal line indicates the median accuracy. The errors were sampled from the results for all wavelengths for the multiple-wavelength models. Note that the expected classification accuracy using a random classifier for a binary classification task such as the one at hand is 50\%.

Having the linear layer programmed using SAM results in a median classification accuracy of 62.5\%, indicating that its poor modeling accuracy has severe practical implications. While the addition of the crosstalk terms in SAM+XT improves the median accuracy to 100\%, more than a quarter of the results have sub-80\% accuracy. In contrast, more than 75\% of all error realizations achieved perfect classification for NN-SW, showcasing how the improved predictions of the data-driven model can make a difference in a practical test case.

Switching over to the multiple-wavelength models, using NN-$\lambda$R for the linear layer results in performance similar to that of SAM-XT. On the other hand, all of the remaining multiple-wavelength models achieve perfect classification for more than 75\% of error realizations. Focusing on the lower whiskers, NN-$\lambda$S, NN-$\lambda$G and TCNN achieve more than 87.5\% accuracy while TCNN-100$\lambda$  achieves 75\% accuracy for more than 90\% of the cases. Overall, the results show that all multiple-wavelength models apart from NN-$\lambda$R can be used for solving this task with high accuracy, but the task having only 8 different input-output pairs for evaluation makes it hard to differentiate between the performances achieved using the NN models.

In order to better quantify the differences between using the NN models for programming a chip for OMM, the same NN architecture was used to solve a regression task where the 3 inputs are $X_1$, $X_2$ and a bias input, and the desired output $Y$ is the probability density function of a 2-D Gaussian distribution with uncorrelated inputs, given in (\ref{eqn_toy_reg}).

\begin{equation}
\label{eqn_toy_reg}
G(X_1,X_2) = \frac{1}{2\pi \sigma_1 \sigma_2}\exp\left( -\frac{(X_1-\mu_1)^2}{2\sigma_1^2} -\frac{(X_2-\mu_2)^2}{2\sigma_2^2}\right)
\end{equation}

\begin{figure}[!t]
\centering
\includegraphics[scale=0.15]{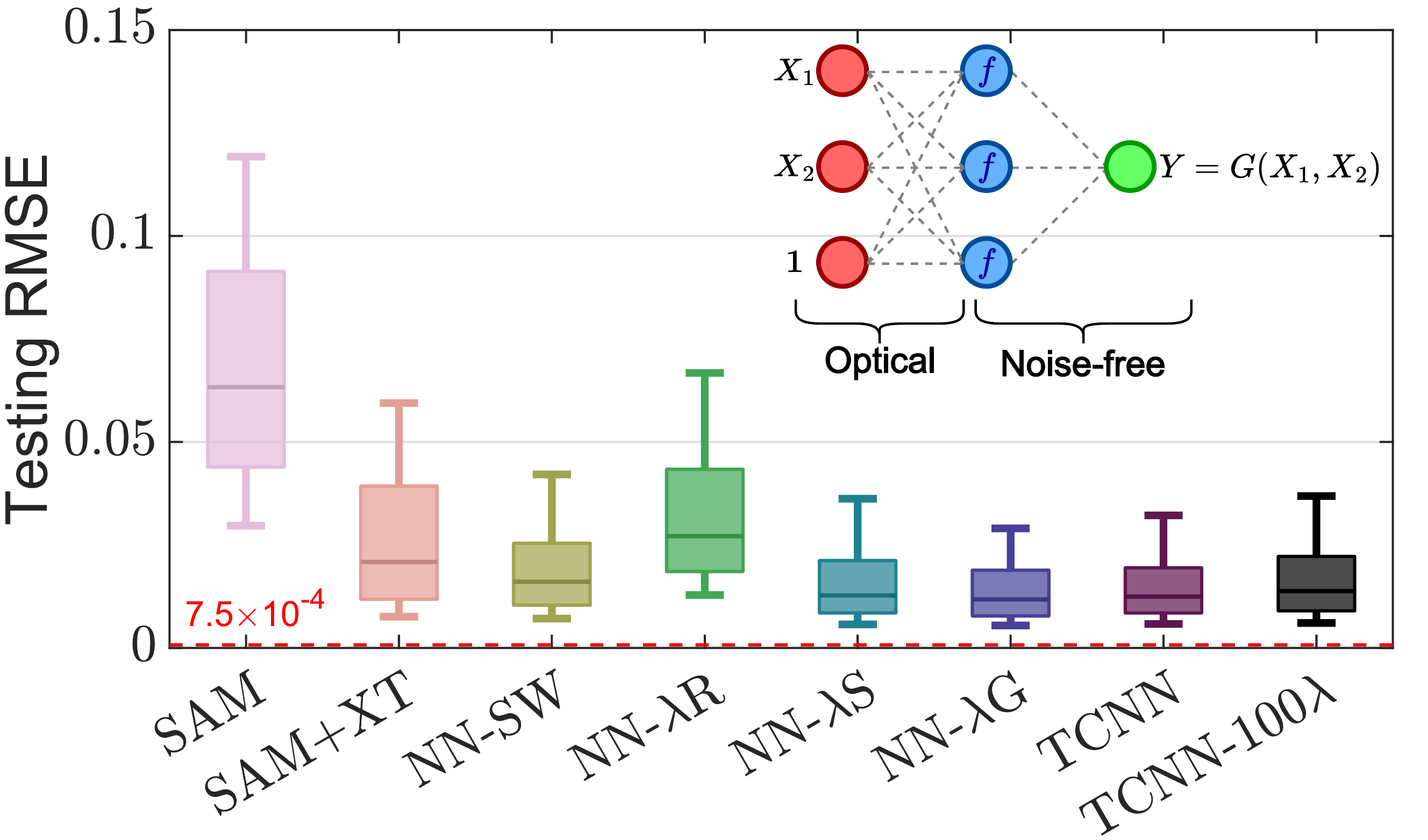}
\caption{Testing RMSE on the 2-D Gaussian function regression task with noise added on the linear layer weights according to the prediction error distributions for the models. Boxes show the $25^{th}$ and $75^{th}$ percentiles while the whiskers show the $10^{th}$ and $90^{th}$ percentiles for 2000 noise realizations. The dark horizontal lines indicate the median accuracies and the red striped line shows the performance of the noise-free NN. Inset: architecture of the NN.}
\label{fig_toy_reg}
\end{figure}

A zero-mean unit-variance distribution i.e. $\mu_1=\mu_2=0$ and $\sigma_1=\sigma_2=1$ was chosen for the regression task. When generating the datasets, 2000 inputs pairs $(X_1,X_2)$ were sampled from two independent uniform random distributions between -1 and +1, which were then split into a training and testing set with 1600 and 400 datapoints, respectively. The noise-free NN achieves a testing RMSE of around $7.5\times 10^{-4}$ after training. Noise was introduced to the NN after training to numerically simulate the performances of the models as described for the previous task and the models were evaluated using the RMSE on the testing set, shown in Fig. \ref{fig_toy_reg}. Boxes show the $25^{th}$ and $75^{th}$ percentiles while the whiskers show the $10^{th}$ and $90^{th}$ percentiles for 2000 noise realizations. The dark horizontal lines indicate the median testing RMSEs and the red striped line shows the performance of the NN before the noise was introduced to the weights of the first linear layer.

The results for the regression task confirm the previous findings: the NN models perform better than the simple analytical models with the exception of SAM+XT performing slightly better than NN-$\lambda$R. While the rest of the models perform similarly, the highest median RMSE was obtained for FNN-SW, followed in decreasing order by TCNN-100$\lambda$, NN-$\lambda$S, TCNN and NN-$\lambda$G. This observation also applies to the other percentiles shown in the graph and is in line with the modeling RMSEs presented earlier.


\section{Conclusion}
We present simple physics-based models and data-driven NN models for OMM with the MZI mesh architecture. Based on our experiments with a fabricated PIC for 3-by-3 matrix-vector multiplication, we conclude that the data-driven models improve the RMSE for the predicted weights from $>1.5$ dB to $0.87$ dB, even when the analytical models include more complex effects such as thermal crosstalk.

The NN models are also capable of achieving modeling errors $<0.80$ dB for 10 adjacent spectral bands within the C-band. In addition, we show that the TCNN architecture can model the matrix weights for up to 100 wavelengths with high accuracy. The multiple-wavelength models achieve promising results for future applications spanning multiple spectral bands in order to accelerate multiple independent tasks by using the same PIC for OMM.

Our results show that accurate models for MZI mesh-based OMM can be obtained using the data-driven approach even under fabrication tolerances and deterministic thermal crosstalk, which are more challenging to avoid as more MZIs are placed in the same chip area. Furthermore, our simulations show that using the NN-based models for programming the PIC for OMM may provide a substantial boost in performance when implementing an optical NN for solving a ML task. While the experiments were performed on a single PIC, the modeling strategies can also be used for PICs where more complex MZI-meshes are implemented.

\appendices

\section{Wavelength Dependence of Matrix Weights}

There are multiple reasons why the matrix weights implemented by a MZI mesh are wavelength dependent, such as the use of directional couplers with wavelength dependent coupling ratios and components with wavelength dependent losses. However, none of these factors should affect the modeling performance of NN-$\lambda$R for our chip as MMIs were used instead of directional couplers for the MZIs and the changes in the losses are compensated for by the rescaling terms $b(\lambda)$. Nevertheless, Fig. \ref{fig_rmse_lambda} shows that the prediction error increases as we move further away from the NN training wavelength $\lambda_6$. In order to make sense of this phenomenon, let us revisit the simple analytical expression for the ratio of the input and output power for a single MZI with 50:50 coupling and infinite extinction ratio.

\begin{equation}
\label{eqn_single_mzi}
w = \frac{P^{out}}{P^{in}} = \alpha \cos^2 \left( \frac{\Delta \Phi}{2} \right)
\end{equation}

$\alpha$ is the optical loss and $\Delta \Phi$ is the total phase difference between the two arms of the MZI, which can be expressed as a function of the heater voltage $v$ and the wavelength.

\begin{equation}
\label{eqn_del_phi}
\Delta \Phi = \frac{2\pi}{\lambda}(\Lambda_0 + \Lambda_2 v^2)
\end{equation}

Note that $\Lambda_0$ is the optical path difference between the arms when $v=0$ and $\Lambda_2$ is the power to optical path difference conversion ratio. Using the notation introduced for SAM, the phase coefficients can be expressed as:

\begin{equation}
\label{eqn_phi0_phi2}
\begin{split}
\phi^{(0)} = \frac{2\pi \Lambda_0}{\lambda} \\
\phi^{(2)} = \frac{2\pi \Lambda_2}{\lambda}
\end{split}
\end{equation}

Focusing on $\phi^{(2)}$, results obtained for SAM show that its value is close to $1 \mathrm{V}^{-2}$. As the wavelength changes by $<1.5\%$ from 1.55 $\mathrm{\mu m}$ for the spectral interval considered in this work, we can approximate the wavelength dependence of $\phi^{(2)}$ by a line, whose slope $s$ at $\lambda=1.55 \mathrm{\mu m}$ is given in (\ref{eqn_slope}).

\begin{equation}
\label{eqn_slope}
s = \frac{d\phi^{(2)}}{d\lambda} = -\frac{\phi^{(2)}}{\lambda} \approx -0.65 \mathrm{\mu m^{-1}}  \mathrm{V}^{-2} 
\end{equation}

\begin{figure}[!t]
\centering
\includegraphics[scale=0.15]{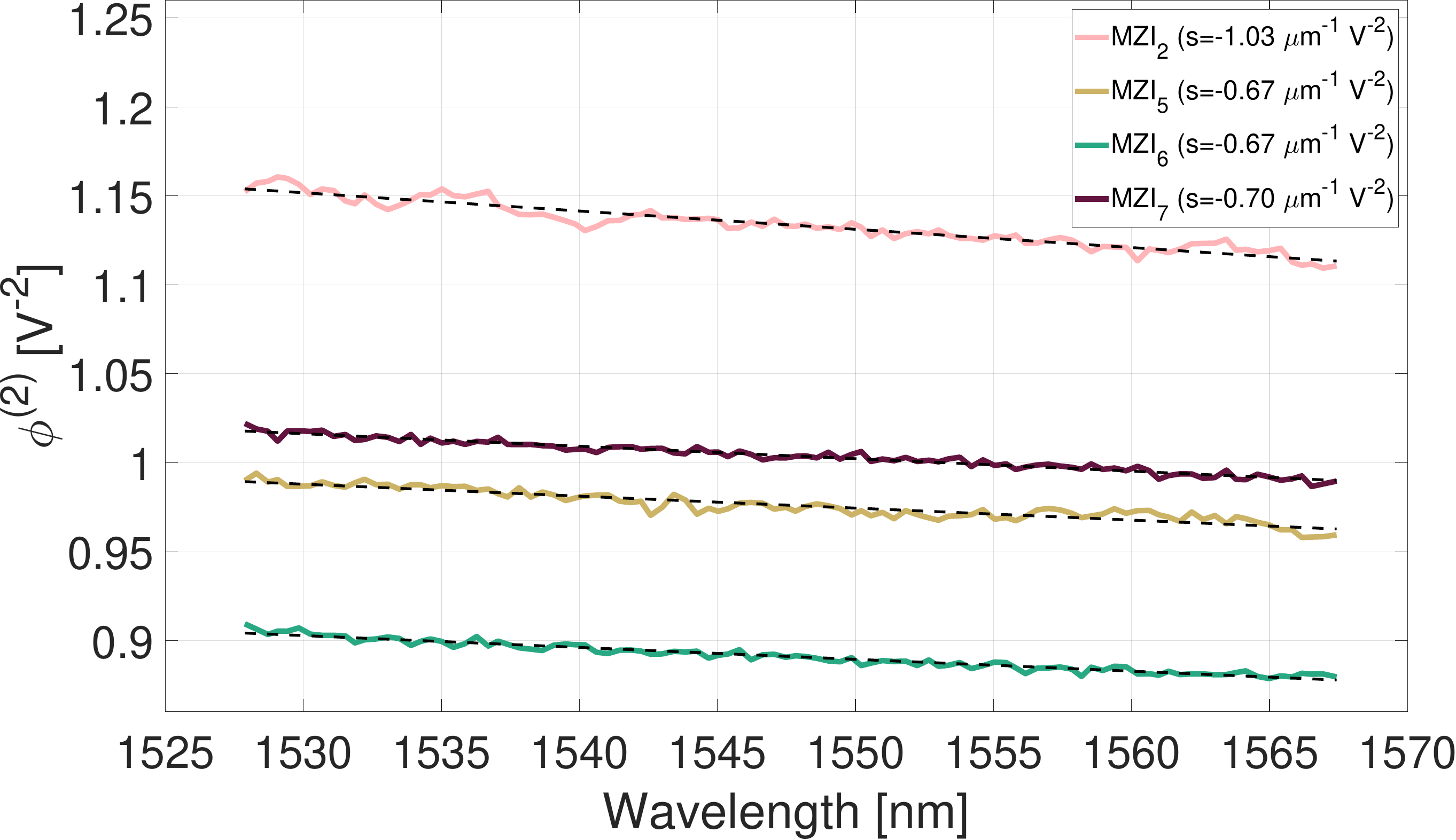}
\caption{Phase coefficient $\phi^{(2)}$ for 4 MZIs obtained by fitting SAM at 100 wavelengths. The dashed black lines are the lines of best fit, whose slopes $s$ are given in the legend.}
\label{fig_phi2_lambda}
\end{figure}

In order to experimentally validate the wavelength dependence of $\phi^{(2)}$, an individual SAM was fit to the measured matrix weights at each of the 100 wavelengths. The resulting values of $\phi^{(2)}$ for 4 of the MZIs are shown in Fig. \ref{fig_phi2_lambda}. The lines of best fit were calculated for the 4 MZIs and their slopes are shown in the legend, which are close to the approximate analytical slope  of $-0.65 \mathrm{\mu m^{-1}} \mathrm{V}^{-2}$. Note that the approximate slope was calculated assuming that $\phi^{(2)} = 1 \mathrm{V}^{-2}$ and the lines should be steeper for $\phi^{(2)} > 1$.

We have shown that $\phi^{(2)}$ is wavelength dependent, indicating that the nature of the relation between the voltages and the weights depend on the wavelength as well. Moreover, $\phi^{(2)}$ decreases almost linearly as the wavelength increases, which explains why training a NN for modeling the PIC at a wavelength further away from the testing wavelength would result in a larger prediction error even when wavelength dependent losses are accounted for, as shown in Fig. \ref{fig_rmse_lambda}.

\section*{Acknowledgment}
This work has received funding by Villum Foundations, Villum YI, OPTIC-AI, grant n. 29344, and ERC CoG FRECOM, grant n. 771878 and National Natural Science Foundation of China n. 62205114, the Key Research and Development Program of Hubei Province, grant n. 2022BAA001.

\newpage

\clearpage

\newpage

\bibliographystyle{IEEEtran}

\bibliography{bibtex/bib/jlt_ref}







\end{document}